\documentclass[final,5p,times]{elsarticle}
\usepackage{pdfpages}
\pdfoutput=1

\usepackage{times, lineno}
\usepackage{epsfig}
\usepackage{graphicx}
\usepackage{amsmath}
\usepackage{amssymb}
\usepackage[bf]{caption}
\usepackage{cs}
\usepackage{mathsymb}
\usepackage{textcomp}
\usepackage{verbatim}
\usepackage{subfigure}
\usepackage{url}
\usepackage{graphicx}
\usepackage{multirow}
\usepackage[super]{nth}
\usepackage{colortbl}
\graphicspath{{./}{./Fig/}{./Figs/}{./Fig/eps/}{./Fig/pdf/}}
\DeclareGraphicsExtensions{.pdf,.png,.jpg}

\usepackage{algorithm2e}

\let\savedalgorithm\algorithm
\let\savedendalgorithm\endalgorithm
\newenvironment{algorithmic}{%
\savedalgorithm
}{%
\savedendalgorithm
}

\usepackage{algorithm}

\def\cone{{\ding{172}}}
\def\ctwo{{\ding{173}}}
\def\cthree{{\ding{174}}}

\def\bPbar{{\bar{\bP}}}

\def\bxi{{\bx_i}}
\def\bxj{{\bx_j}}
\def\bxip{{\bx_i^{+}}}
\def\bxijm{{\bx_{i,j}^{-}}}
\def\mprime{{m^{\prime}}}

\def\bdip{{\bd_{i}^{+}}}
\def\bdjm{{\bd_{j}^{-}}}
\def\bdjbm{{\bd_{j}^{-}}}

\def\bPast{{{\boldsymbol P}^{\ast}}}

\def\Integer{\mathbb{Z}^+}

\def\CMCstruct{{\rm CMC$^{\,\rm top}$}\xspace}
\def\CMCtriplet{{\rm CMC$^{\,\rm triplet}$}\xspace}

\newcommand{\ie}{\emph{i,e.,}}

\newcommand{\CScomment}[1]{}

\journal{}

\begin{document}

\begin{frontmatter}

\title{Structured learning of metric ensembles with application to person re-identification}

\author[label1]{Sakrapee Paisitkriangkrai}
\author[label1,label2]{Lin Wu }
\author[label1,label2]{Chunhua Shen \corref{cor1}}
\ead{chhshen@gmail.com}
\author[label1,label2]{Anton van den Hengel}

\cortext[cor1]{Corresponding author.}

\address[label1]{School of Computer Science, The University of Adelaide, Australia}
\address[label2]{Australian Centre for Robotic Vision,  Australia}

\begin{abstract}
Matching individuals across non-overlapping camera networks, known as person re-identification, is a fundamentally challenging problem due to the large visual appearance changes caused by variations of viewpoints, lighting, and occlusion. Approaches in literature can be categorized into two streams: The first stream is to develop reliable features against realistic conditions by combining several visual features in a pre-defined way; the second stream is to learn a metric from training data to ensure strong inter-class differences and intra-class similarities.  However, seeking an optimal combination of visual features which is generic yet adaptive to different benchmarks is an unsolved problem, and metric learning models easily get over-fitted due to the scarcity of training data in person re-identification.
In this paper, we propose two effective structured learning based approaches which explore the adaptive effects of visual features in recognizing persons in different benchmark data sets.
Our framework is built on the basis of multiple low-level visual features with an optimal ensemble of their metrics.
We formulate two optimization algorithms, \CMCtriplet and \CMCstruct, which directly optimize evaluation measures commonly used in person re-identification, also known as the Cumulative Matching Characteristic (CMC) curve. The more standard \CMCtriplet formulation works on the triplet information by maximizing the relative distance between a matched pair and a mismatched pair in each triplet unit.
The \CMCstruct formulation, modeled on a structured learning of maximizing the correct identification among top candidates, is demonstrated to be more beneficial to person re-identification by directly optimizing an objective closer to the actual testing criteria.
The combination of these factors leads to a person re-identification system which outperforms most existing algorithms.
More importantly, we advance state-of-the-art results by improving the rank-$1$ recognition rates from $40\%$ to $61\%$ on the iLIDS benchmark, $16\%$ to $22\%$ on the PRID2011 benchmark, $43\%$ to $50\%$ on the VIPeR benchmark, $34\%$ to $55\%$ on the CUHK01 benchmark and $21\%$ to $68\%$ on the CUHK03 benchmark.
\end{abstract}


\end{frontmatter}

\section{Introduction}

The task of person re-identification (re-id) is to match pedestrian images observed from different and disjoint camera views.
Despite extensive research efforts in re-id \cite{Gheissari2006Person,Roth2014Mahalanobis,Zhao2014Learning,Zheng2011Person,Xiong2014Person,Zhao2013Unsupervised,Pedagadi2013Local,Li2013Learning,Zhao2013SalMatch}, the problem itself is still a very challenging task due to (a) large variation in visual appearance (person's appearance often undergoes large variations
across different camera views); (b) significant changes in human poses at the time the image was captured;
(c) large amount of illumination changes, background clutter and occlusions; d) relatively low resolution and the different placement of the cameras. Moreover, the problem becomes increasingly difficult when there are high variations in pose, camera viewpoints, and illumination, etc.

To address these challenges, existing research has concentrated on the development of sophisticated and robust features to describe visual appearance under significant changes. Most of them use appearance-based features that are viewpoint invariant such as color and texture descriptors \cite{Farenzena2010Person,Gheissari2006Person,Gray2008Viewpoint,Wang2007Shape,WhatFeatures}.
However, the system that relies heavily on one specific type
of visual cues, e.g., color, texture or shape, would not be practical and powerful enough to discriminate individuals with
similar visual appearance. Some studies have tried to address the above problem by seeking a combination of robust and distinctive feature representation of person's appearance, ranging from color histogram \cite{Gray2008Viewpoint}, spatial co-occurrence representation \cite{Wang2007Shape}, LBP \cite{Xiong2014Person}, to color SIFT \cite{Zhao2013Unsupervised}.
The basic idea of exploiting multiple visual features is to build an ensemble of metrics (distance functions),
in which each distance function is learned using a single feature and the final distance is calculated from a weighted
sum of these distance functions \cite{Farenzena2010Person,Xiong2014Person,Zhao2013Unsupervised}. These works often pre-define distance weights, which need to be re-estimated beforehand for different data sets. However, such a pre-defined principle has some drawbacks.
\begin{itemize}
  \item Different real-world re-id scenarios can have very different characteristics, e.g., variation in view angle, lighting and occlusion.
Simply combining multiple distance functions using pre-determined weights may be undesirable
as highly discriminative features in one environment might become irrelevant in another environment.
\item The effectiveness of distance learning heavily relies on the quality of the feature selected,
  and such selection requires some domain knowledge and expertise.
\item Given that certain features are determined to be more reliable than others under a certain condition,
  applying a standard distance measure for each individual match is undesirable
  as it treats all features equally without differentiation on features.
\end{itemize}
In these ends, it necessarily demands a principled approach that
is able to automatically select and learn weights for diverse metrics, meanwhile generic yet adaptive to different scenarios.

Person re-identification problem can also be cast as a learning problem in which either metrics or discriminative models are learned \cite{Chopra2005Learning,Davis2007Information,Kedem2012Nonlinear,Kostinger2012Large,Weinberger2008Fast,Wu2011Optimizing,Xiong2014Person,Li2013Learning,Weinberger2006Distance,PCCA,Zheng2013PAMI}, which  typically learn a distance measure by minimizing intra-class distance and maximizing inter-class distance simultaneously. Thereby, they require sufficient labeled training data from each class \footnote{Images of each person in a training set form a class.} and most of them also require new training data when camera settings change. Nonetheless, in person re-id benchmark, available training data is relatively scarce, and thus inherently undersampled for building a representative class distribution. This intrinsic characteristic of person re-id problem makes metric learning pipelines easily overfitted and unable to be applicable in small image sets.

\begin{figure*}[t]
    \centering
    \includegraphics[width=0.9\textwidth]{{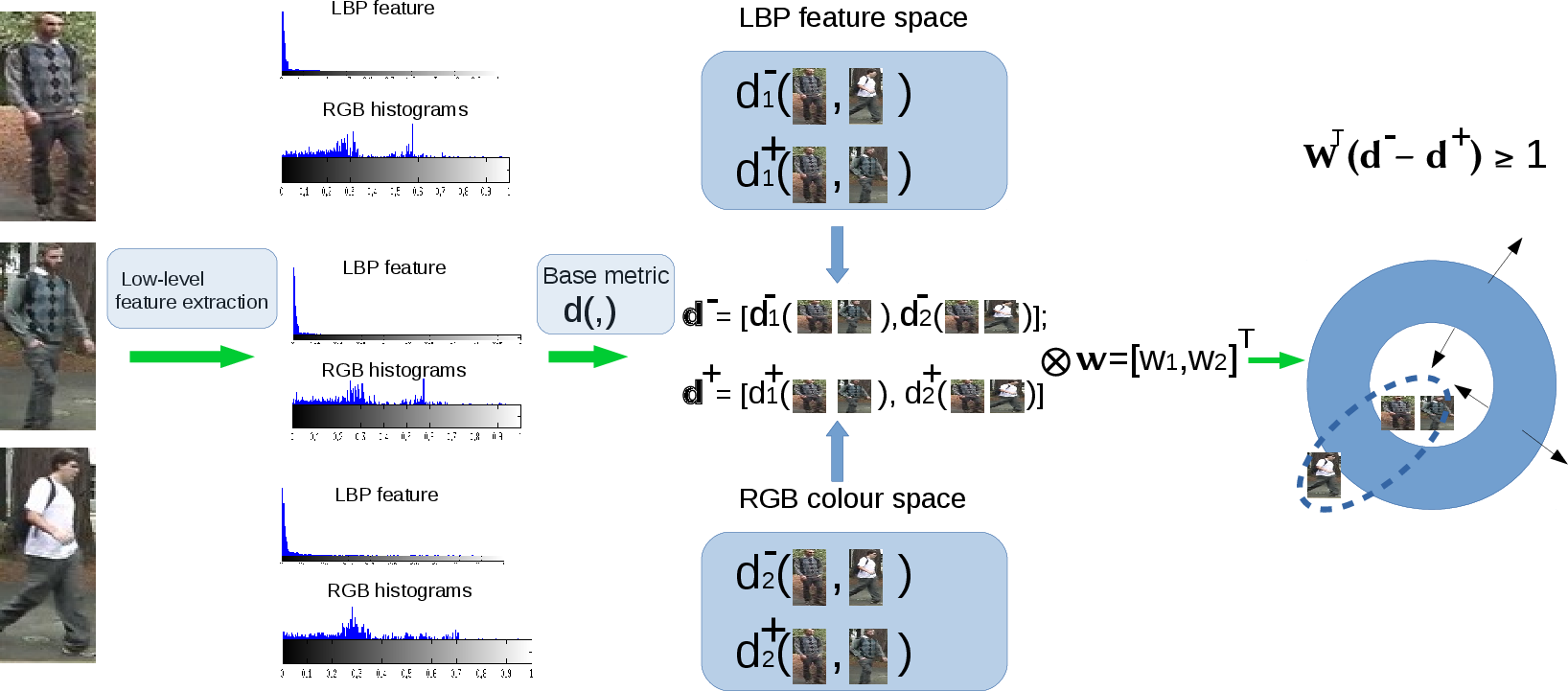}}
    \caption{An illustration of our triplet-based framework. Given a set of triplets, our model is trained to fully exploit low-level features by adaptively learning weights for their distance functions, so as to maximize the relative distance between the matched pair and the mismatched pair for each triplet.  }\label{fig:framework}
\end{figure*}

To combat above difficulties simultaneously, in this paper, we introduce two structured learning based approaches to person re-id by learning weights of distance functions for low-level features. The first approach, \CMCtriplet, optimizes the relative distance using the triplet units, each of which contains three person images, i.e., one person with a matched reference and a mismatched reference.
Treating these triplet units as input, we formulate a large margin framework with triplet loss where the relative distance between the matched pair and the mismatched pair tends to be maximized.  An illustration of \CMCtriplet is shown in Fig. \ref{fig:framework}.
This triplet based model is more natural for person re-id mainly because the intra-class and inter-class variation may
vary significantly for different classes,
making it inappropriate to require the distance between a matched/mismatched pair to fall within an absolute range \cite{Zheng2013PAMI}.
Also, training images in person re-id are relatively scarce, whereas the triplet-based training model is to make comparison between any two data points rather than comparison between any data distribution boundaries or among clusters of data. This thus alleviates the over-fitting problem in person re-id given undersampled data.
The second approach, \CMCstruct, is developed to maximize the average rank-$k$ recognition rate, in which $k$ is chosen to be small, e.g., $k < 10$. Setting the value of $k$ to be small is crucial for many real-world applications since most surveillance operators typically inspect only the first ten or twenty items retrieved. Thus, we directly optimize the testing performance measure commonly used in CMC curve, i.e., the recognition rate at rank-$k$ by using structured learning.

The main contributions of this paper are three-fold:
\begin{itemize}
\item We propose two principled approaches, \CMCtriplet and \CMCstruct, to build an ensemble of person re-id metrics. The standard approach \CMCtriplet is developed based on triplet information, which is more tolerant to large intra and inter-class variations, and alleviate the over-fitting problem. The second approach of \CMCstruct directly optimizes an objective closer to the testing criteria by maximizing the correctness among top $k$ matches using structured learning, which is empirically demonstrated to be more beneficial to improving recognition rates.
\item We perform feature quantification by exploring the effects of diverse feature descriptors in recognizing persons in different benchmarks. An ensemble of metrics is formulated into a late fusion paradigm where a set of weights corresponding to visual features are automatically learned. This late fusion scheme is empirically studied to be superior to various early fusions on visual features.
\item Extensive experiments are carried out to demonstrate that by building an ensemble of person re-id metrics learned from different visual features, notable improvement on rank-$1$ recognition rate can be obtained.  In addition, our ensemble approaches are highly flexible and can be combined with  linear and non-linear metrics. For  non-linear base metrics, we extend our approaches to be tractable and suitable to  large-scale benchmark data sets by approximating the kernel learning.
\end{itemize}

\section{Related Work}\label{sec:related}

Many person re-id approaches are proposed to seek robust and discriminative features  such that they can be used to describe the appearance
of the same individual across different camera views under various changes and conditions \cite{Bazzani2012Multiple, Cheng2011Custom,Farenzena2010Person,Gheissari2006Person,Gray2008Viewpoint,Li2014Deep,Wang2007Shape,Zhao2013Unsupervised,Zhao2014Learning}.
For instance, Bazzani \etal represent a person by a global mean color histogram and recurrent local patterns through epitomic analysis \cite{Bazzani2012Multiple}. Farenzena \etal propose the symmetry-driven accumulation of local features (SDALF) which exploits both symmetry and asymmetry, and represents each part of a person by a weighted color histogram, maximally stable color regions and texture information \cite{Farenzena2010Person}. Gray and Tao introduce an ensemble of local features which combines three color channels with $19$ texture channels \cite{Gray2008Viewpoint}. Schwartz and Davis propose a discriminative appearance based model using partial least squares where multiple visual features: texture, gradient and color features are combined \cite{Schwartz2009Learning}. Zhao \etal combine dcolorSIFT with unsupervised salience learning to improve its discriminative power in person re-id \cite{Zhao2013Unsupervised} (eSDC), and further integrate both salience matching and patch matching into a unified RankSVM framework (SalMatch \cite{Zhao2013SalMatch}). They also propose mid-level filters (Mid\-Le\-ve\-l) for person re-identification by exploring the partial area under the ROC curve (pAUC) score \cite{Zhao2014Learning}. Lisanti \etal \cite{Giuseppe2015PAMI} leverage  low-level feature descriptors to approximate the appearance variants in order to discriminate individuals by using sparse linear reconstruction model (ISR).

Another line to approach the problem of matching people across cameras is  to essentially formalize person re-id as a supervised metric/distance learning where a projection matrix is sought out so that the projected Mahalanobis-like distance is small when feature vectors represent the same person and large otherwise. Along this line, a large number of metric learning and ranking algorithms have been proposed \cite{Chopra2005Learning,Guillaumin2009Isthatyou,Davis2007Information,Pedagadi2013Local,Kedem2012Nonlinear,Kostinger2012Large,
Weinberger2006Distance,Weinberger2008Fast,Wu2011Optimizing,Xiong2014Person,Li2013Learning,PCCA,Zheng2013PAMI}.
Among these Mahalanobis distance learning algorithms, the Large Margin Nearest Neighbor Learning (LMNN) \cite{Weinberger2006Distance}, Information Theoretic Metric Learning (ITML) \cite{Davis2007Information}, and Logistic Discriminant Metric Learning (LDML) \cite{Guillaumin2009Isthatyou} are three representative methods.

Metric learning methods can be applied to person re-id to achieve the state of the art results \cite{Hirzer2012Person,Kostinger2012Large,Li2013Learning} .  In particular,
Koestinger \etal propose the large-scale metric learning from equivalence constraint (KISSME) which considers a log likelihood ratio test of two Gaussian distributions \cite{Kostinger2012Large}.
Li \etal propose the learning of locally adaptive decision functions (LADF) \cite{Li2013Learning}.
An alternative approach is to use a logistic function to approximate the hinge loss so that the global optimum still can be achieved by iteratively gradient search along the projection matrix as in PCCA \cite{PCCA}, PRDC \cite{Zheng2013PAMI} and Cross-view Quadratic Discriminant Analysis (XQDA) \cite{XQDA}. However, these methods are prone to over-fitting and can result in poor classification performance due to large variations among samples. We approach this problem by proposing a more generic algorithm which is developed based on triplet information so as to generate more constraints for distance learning, and thus mitigate the over-fitting issue. Prosser \etal use pairs of similar and dissimilar images and train the ensemble RankSVM such that the true match gets the highest rank \cite{Prosser2010Person}. However, the model of RankSVM needs to determine the weight between the margin function and the ranking error cost function, which is computationally costly. Wu \etal applies the Metric Learning to Rank (MLR) method of \cite{McFee2010Metric} to person re-id \cite{Wu2011Optimizing}.

Although a large number of existing algorithms have exploited state-of-the-art visual features and advanced metric learning algorithms, we observe that the best obtained overall performance on commonly evaluated person re-id benchmarks, e.g., iLIDS and VIPeR, is still far from the performance needed for most real-world surveillance applications. The goal of this paper is to show how one can carefully design a standard, generalized person re-id framework to achieve state-of-the-art results on person re-id testing data sets.
Preliminary results of this work were published in \cite{paul2015ensemble}.

\section{Notations and Problem Definition}\label{sec:notation}

Throughout the paper, we denote vectors with bolded font and matrices with capital letters. We focus on the single-shot modality where there is a single exemplar for each person in the gallery and one exemplar for each person in the probe set.
We represent a set of training samples by $\left\{ (\bxi, \bxip) \right\}_{i=1}^m$
where $\bx_i \in {\Real}^D$ represents a training example from one camera, $\bxip$ is the corresponding image of the same person from a different camera, and  $m$ is the number of persons in the training data.
Then, a set of triplets for each sample $\bxi$ can be generated as
$\left\{(\bxi, \bxip, \bxijm) \right\}$ for $i = 1,\cdots,m$ and $i \neq j$.
Here we introduce $\bxijm \in \cX_i^{-}$ where $\cX_i^{-}$ denotes a subset of images of persons with a different identity to $\bxi$ from a different camera view.
We assume that there exist a set of distance functions $d_t(\cdot,\cdot)$ which calculate the distance between two given inputs.
Our goal is to learn a weighted distance function:
$d(\cdot,\cdot) = \sum_{t=1}^T w_t d_t(\cdot,\cdot)$,
such that the distance between $\bxi$ and $\bxip$  is smaller than the distance between
$\bxi$ and any $\bxijm$.
A good distance function can facilitate the cumulative matching characteristic (CMC)
curve  to approach one faster.
\section{Our Approach}\label{sec:approach}

In this section, two structured learning based approaches are presented to learn an ensemble of base metrics. We then discuss base metrics that will be used in our experiment as well as a strategy of approximating non-linear metric learning for the case of large-sized person re-id data set.

\subsection{Ensemble of base metrics}\label{sec:ensemble}
We propose two different approaches to learn an ensemble of base metrics in an attempt to rank the possible candidates such that the highest ranked candidate is the correct match for a query image.
In our setting, we consider the most commonly used performance measure for evaluating person re-id, cumulative matching characteristic (CMC) curve \cite{Gray2007Evaluating}, which represents results of an identification task by plotting the probability of correct identification against the number of candidates returned.
A better person re-id algorithm can make the CMC curve approach one faster.
Achieving the best rank-$1$ recognition rate is the ultimate goal \cite{Zhao2014Learning} in many real-world surveillance applications because most users are more likely to consider the first a few returned candidates.  Thus, our goal is to improve the recognition rate among the $k$ best candidates with a minimized $k$, e.g., $k < 20$.
The first approach, \CMCtriplet, aims at minimizing the number of returned list of candidates in order to achieve a perfect identification, i.e., minimizing $k$ such that the rank-$k$ recognition rate is equal to one. The second approach, \CMCstruct, optimizes the probability that any of these $k$ best matches are correct.

\subsection{Relative distance based approach (\CMCtriplet)}
The \CMCtriplet approach aims to optimize the rank-$k$ recognition rate of the candidate list for a given probe image. To do this, we propose to learn an ensemble of distance
functions based on relative comparison of triplets \cite{Schultz2004Learning}.
The relative distance relationship is reflected by a set of triplet units $\left\{(\bxi, \bxip, \bxijm) \right\}_{i,j}$. For an image $\bxi$ of a person, we wish to learn a re-id model to successfully identify another image $ \bxip$ of the same person captured by another camera view. For a triplet $\left\{ (\bxi, \bxip, \bxijm) \right\}_{i,j}$, we assume a distance satisfying $d(\bxi,\bxijm) > d(\bxi,\bxip)$, where $\bxijm$ is an image of any other person expect $\bxi$. Inspired by the large margin framework with the hinge loss, a large margin is expected to exit between positive pairs and negative pairs, that is $d(\bxi,\bxijm) \geq 1 + d(\bxi,\bxip)$. Considering this margin condition cannot be satisfied by all triplets, we introduce a slack variable to enable soft margin. Finally, by generalizing this inequality to the entire training set, the primal problem that we want to optimize is
\begin{align}\label{EQ:svm}
        \min_{ \bw, \bslack }   \;
        & \frac{1}{2} \| \bw  \|_{2}^{2} + \nu \, \frac{1}{m(m-1)} \;
                                \sum_{i=1}^m \sum_{j=1}^{m-1} \xi_{ij}   \\ \notag
        \st \; & \bw^\T ( \bdjm - \bdip ) \geq 1 - \xi_{ij}; \bw \geq 0; \; \bslack \geq 0.
\end{align}
Here $\bw$ is the vector of the base metric weights, $m$ denotes the total number of identities in the training set. $\nu > 0$ is the regularization parameter and
$\bdjm$ = $[d_1(\bxi,\bxijm),$ $\cdots,$ $d_t(\bxi,\bxijm)]$, $\bdip$ = $[d_1(\bxi,\bxip),$ $\cdots,$ $d_t(\bxi,\bxip)]$ and
$\{d_1(\cdot,\cdot)$, $\cdots$, $d_t(\cdot,\cdot)\}$ represent a set of base metrics. The regularization term $\| \bw \|_{2}^2$ is introduced to avoid the trivial solution of arbitrarily large $\bw$. It can be seen that the number of constraints in \eqref{EQ:svm} is quadratic in terms of the size of training examples, and directly solving \eqref{EQ:svm} using off-the-shelf optimization toolboxes is intractable. To address this issue, we present an equivalent reformulation of \eqref{EQ:svm}, which can be efficiently solved in a linear runtime using the cutting-plane algorithm. We first reformulate \eqref{EQ:svm} to be:
\begin{align}\label{EQ:svm2}
        \min_{ \bw, \xi }   \;
        &
         \frac{1}{2} \| \bw  \|_{2}^{2} + \nu \,  \xi   \\ \notag
        \st \; &
        \frac{1}{m(m-1)} \bw^\T \Bigl[ \sum_{i=1}^m \sum_{j=1}^{m-1}
                ( \bdjm - \bdip ) \Bigr] \geq 1 - \xi; \bw \geq 0; \; \slack \geq 0.
\end{align}
This new formulation has a single slack variable and later on we show  the cutting-plane method can be applied to solve it.

\subsection{Top recognition at rank-$k$ (\CMCstruct)} In \CMCtriplet, we set up the hypothesis that for any triplet, images belonging to the same identity should be closer than images from different identities. Our second formulation is motivated by the fact that person re-id users often concentrate on only the first few retrieved matches. Bearing this in mind, we propose \CMCstruct with an objective of maximizing the correct identification among the top $k$ best candidates. Specifically, we directly optimize the performance measure indicated by the CMC curve (recognition rate at rank-$k$) by taking advantage of structured learning framework \cite{Joachims2005Support, Narasimhan2013Structural}.
The main difference between our work and \cite{Narasimhan2013Structural}
lies in the fact that \cite{Narasimhan2013Structural} attempts to rank all positive samples before a subset of negative samples while our works attempt to rank a pair of the same individual above a pair of different individuals. Given a training identity $\bxi$ and its correct match $\bxip$ from a different view, the relative ordering of all matching candidates in different view can be represented via a vector $\bp \in \Real^\mprime$, in which $p_j$ is $0$ if $\bxip$ is ranked {\em above} $\bxijm$ and $1$ if $\bxip$ is ranked {\em below} $\bxijm$. Here $\mprime$ indicates the total number of identities who has a different identity to $\bxi$. Since there exists only one match of the same identity, $\mprime$ is equal to $m-1$. We can generalize this idea to the entire training set and represent the relative ordering via a matrix $\bP \in \left\{ 0, 1 \right\}^{m \times \mprime}$ as follows:
\begin{align} \label{EQ:piij}
        p_{ij}  =
            \begin{cases}
                0  \quad& \text{if} \; \bxip \; \text{is ranked above} \; \bxijm  \\
                1  \quad& \text{otherwise.}
            \end{cases}
\end{align}
The correct relative ordering of $\bP$ can be defined as $\bPast$ where $p^{\ast}_{ij} = 0, \forall i, j$. The loss among the top $k$ candidates between $\bPast$ and an arbitrary ordering $\bP$ can be formulated as,
\begin{align}
    \label{EQ:delta}
        \Delta (\bPast, \bP) = \frac{1}{m \cdot k} \; \sum_{i=1}^m
           \sum_{j=1}^k p_{i,j},
\end{align}
where $j$ denotes the index of the retrieved candidates ranked in the $j$-th position among all top $k$ best candidates.
We define the joint reward score, $\psi$, of the form:
\begin{align}
    \label{EQ:featmap}
        \psi (\bS, \bP) = \frac{1}{m \cdot k}
            \sum_{i=1}^m \sum_{j=1}^k
            (1 - p_{ij}) (\bdjm - \bdip),
\end{align}
where $\bS$ represent a set of triplets generated from the training data,
$\bdjm$ = $[d_1(\bxi,\bxijm),$ $\cdots,$ $d_t(\bxi,\bxijm)]$ and
$\bdip$ = $[d_1(\bxi,\bxip),$ $\cdots,$ $d_t(\bxi,\bxip)]$.
The intuition of Eq. \eqref{EQ:featmap} is to establish a structure consistency between relative distance margins and corresponding rankings in $\bS$. Specifically, the distance margin ($\bdjm - \bdip$) is scaled proportionally to the loss in the resulting ordering ($1-p_{ij}$). This can ensure a high reward score can be computed by Eq. \eqref{EQ:featmap} if postive and negative pairs exhibit large margins and meanwhile
$\bxip$ is always ranked {\em above} $\bxijm$ (\ie $p_{ij}=0, j=1,\dots,k$). Thus, $\psi(\bS, \bP)$ can be seen as a joint reward score to reflect the structure consistency between $\bS$ and $\bP$.

In fact, the choice of $\psi(\bS, \bP)$ is to guarantee that $\bw$, which optimizes
$\bw^\T \psi(\bS, \bP)$, will also produce the distance function
$d(\cdot,\cdot) = \sum_{g=1}^t w_g d_g(\cdot,\cdot)$ that
achieves the optimal average recognition rate among
the top $k$ candidates. The above problem can be summarized as the following convex
optimization problem:
\begin{align}\label{EQ:struct}
    \min_{ \bw , \xi }   \quad
    &
    \frac{1}{2} \| \bw  \|_{2}^{2} + \nu \, \xi    \\ \notag
    \st \; &
    \bw^\T \bigl( \psi(\bS, \bPast) - \psi(\bS, \bP) \bigr)
    \geq \Delta (\bP^{\ast}, \bP) - \xi; \forall \bP \geq 0; \xi \geq 0.
\end{align}

The intuition of Eq \eqref{EQ:struct} is to ensure there is a large margin ($\Delta (\bP^{\ast}, \bP)$) between the reward score from the correct ordering $\psi(\bS, \bPast)$ and any estimated ordering $\psi(\bS, \bP)$ with respect to $\bS$.

\subsubsection{Cutting-plane optimization} The optimization problems in \eqref{EQ:svm2} and \eqref{EQ:struct} involve quadratic program, which is unable to be solved directly. Hence, we resort to cutting-plane method to approximate its solution.
The cutting-plane optimization works by assuming that a small subset of the constraints are sufficient to find an $\epsilon$-approximate solution to the original problem.
The algorithm begins with an empty initial
constraint set and iteratively adds the most violated constraint set.
At each iteration, it computes the solution over the current working set, finds the most violated constraint, and puts it into the working set.
The algorithm continues until no constraint is violated
by more than $\epsilon$. Since the quadratic program is of constant size, the cutting-plane method
converges in a constant number of iterations. We present \CMCstruct in Algorithm~\ref{ALG:cutting}.

In \CMCstruct , the optimization problem for finding the most violated constraint
(Algorithm~\ref{ALG:cutting}, step \ctwo) can be written as,
\begin{align}\label{EQ:violated1}
    \bPbar &= \max_{ \bP }  \Delta (\bP^{\ast}, \bP) -
                \bw^\T \bigl( \psi(\bS, \bPast) - \psi(\bS, \bP) \bigr) \\ \notag
           &= \max_{ \bP }  \Delta (\bPast, \bP) - \frac{1}{mk}
         \sum_{i,j}  p_{ij} \bw^\T (\bdjm - \bdip ) \\ \notag
           &= \max_{ \bP } \sum_{i=1}^m \Bigl(\sum_{j=1}^k p_{i,j} (1-\bw^\T \bd_{i,j}^{\pm})
               - \sum_{j=k+1}^\mprime p_{i,j} \bw^\T \bd_{i,j}^{\pm} \Bigr)
\end{align}
where $\bd_{i,j}^{\pm} = \bdjbm - \bdip$. Since $p_{ij}$ in \eqref{EQ:violated1} is independent, the solution to \eqref{EQ:violated1} can be solved by maximizing over each element $p_{ij}$. Hence, in $\bPbar$ the most violated constraint corresponds to,
\begin{align}
  \bar{p}_{i,j} = \left\{ \begin{array}{ll}
  \b1 \bigl( \bw^\T (\bdjbm - \bdip) \leq 1 \bigr) , &\mbox{{if   }} j \in \{ 1, \cdots, k \} \\
  \b1 \bigl( \bw^\T (\bdjbm - \bdip) \leq 0 \bigr), &\mbox{{otherwise.}}
\end{array} \right. \notag
\end{align}

For \CMCtriplet, one replaces $g(\bS, \bP, \bw)$ in Algorithm~\ref{ALG:cutting} with
$g(\bS, \bw) = 1 - \frac{1}{m(m-1)} \bw^\T \left[ \sum_{i,j} ( \bdjm - \bdip ) \right]$ and repeats the same procedure.

\SetKwInput{KwInit}{Initialize}
\SetKwRepeat{Repeat}{Repeat}{Until}

\begin{algorithm}[t]
\caption{Cutting-plane algorithm for solving  \CMCstruct (\CMCtriplet)
}
\begin{algorithmic}
\footnotesize{
   \KwIn{
     \\1)   A set of base metrics of the same identity and different identities $\{\bdip, \bdjm\}$;
     $
     \;
     $
     \\2)    The regularization parameter, $\nu$;
     $
     \;
     $
     \\3)    The cutting-plane termination threshold, $\epsilon$;
   }

   \KwOut{
      The base metrics' coefficients $\bw$,
    }

\KwInit {
    The working set, $\cC = \varnothing;$
}

$g(\bS, \bP, \bw) = \Delta (\bPast, \bP) - \frac{1}{mk}
         \sum_{i,j}  p_{ij} \bw^\T (\bdjm - \bdip );$

\Repeat{ $ g(\bS, \bP, \bw) \leq \xi + \epsilon $
}{
\cone\ Solve the primal problem using linear SVM,
\begin{flalign}
    \notag
    \min_{ \bw , \xi } \;
    \frac{1}{2} \| \bw  \|_{2}^{2} + \nu \, \xi  \quad
    \st \;
    g(\bS, \bP, \bw) \leq \xi, \forall \bP \in \cC;
\end{flalign}
\ctwo\ Compute the most violated constraint,
\begin{align}
    \notag
    \bPbar = \max_{ \bP }  g(\bS, \bP, \bw);
\end{align}
\cthree\ $\cC \leftarrow \cC \cup  \{ \bPbar \};$
}
} %
\end{algorithmic}
\label{ALG:cutting}
\end{algorithm}

\subsection{Base metrics}\label{sec:metric}
In this paper, two types of base metrics are employed to develop our structured learning algorithms, which are KISS metric learning \cite{Kostinger2012Large} and kernel Local Fisher Discriminant Analysis (kLFDA) \cite{Xiong2014Person}.

\paragraph{KISS metric learning revisit}
The KISS ML method learns a linear mapping by estimating a matrix $\bM$ such that the distance between images of the same individual, $(\bxi - \bxip)^\T \bM (\bxi - \bxip)$, is less than the distance between images of different individuals, $(\bxi - \bxijm)^\T \bM (\bxi - \bxijm)$.
Specifically, it considers two independent generation processes for observed images of similar and dissimilar pairs. From a statistical view, the optimal statistical decision whether a pair ($\bx_{i}, \bx_{j}$) is dissimilar or not can be obtained by a likelihood ratio test.
Thus, the likelihood ratio test between dissimilar pairs and similar pairs is,
\begin{align}
    \label{EQ:KISSME1}
    r(\bxi,\bxj) = \log \frac{ \frac{1}{\sqrt{2 \pi | \Sigma_{\cD} |}} \exp(-\frac{1}{2} \bx_{ij}^\T \Sigma_{\cD}^{-1} \bx_{ij}  )}
                             { \frac{1}{\sqrt{2 \pi | \Sigma_{\cS} |}} \exp(-\frac{1}{2} \bx_{ij}^\T \Sigma_{\cS}^{-1} \bx_{ij}  )} ,
\end{align}
where $\bx_{ij} = \bx_{i} - \bx_{j}$, $\Sigma_{\cD}$ and $\Sigma_{\cS}$ are covariance matrices of dissimilar pairs and similar pairs, respectively. By applying the Bayesian rule and the log-likelehood ratio test, the decision function can be simplied as $f(\bx_{ij}) = \bx_{ij}^T (\Sigma_{\cS}^{-1} - \Sigma_{\cD}^{-1}) \bx_{ij}$,
and the derived distance function between $\bx_{i}$ and $\bx_{j}$ is
\begin{align}
    \label{EQ:KISSME2}
    d(\bxi,\bxj) = (\bx_{i} - \bx_{j})^\T (\Sigma_{\cS}^{-1} - \Sigma_{\cD}^{-1})
                      (\bx_{i} - \bx_{j}).
\end{align}
Hence, learning the distance function corresponds to estimating the covariant matrices $\Sigma_{\cS}$ and $\Sigma_{\cD}$, and we have the Mahalanobis distance matrix $\bM$ as  $\bM=\Sigma_{\cS}^{-1} - \Sigma_{\cD}^{-1}$.
In\cite{Kostinger2012Large}, $\bM$ is obtained by clipping the spectrum of $\bM$ through eigen-analysis to ensure the property of positive semi-definite. This simple algorithm has shown to perform surprisingly well on the person re-id problem \cite{Roth2014Mahalanobis, Li2014Deep}.

\paragraph{kLFDA revisit}
kLFDA is a non-linear extension to LFDA \cite{Pedagadi2013Local} and has demonstrated the state-of-the-art performance on person re-id problem. kLFDA is
a supervised dimensionality reduction method that  uses a kernel trick to handle high dimensional feature vectors  while maximizing the Fisher optimization criteria. A projection matrix $\bM$ can be achieved to maximize the between-class scatter matrix while minimizing the within-class scatter matrix for similar samples using the Fisher discriminant objective.  kLFDA represents the projection matrix with data samples in the kernel space $\phi(\cdot)$.
Once the kernel function is computed, $\kappa(\bx,\bx^\prime) = \phi(\bx)^\T \phi(\bx^\prime)$, we can efficiently learn $\bM$.

\subsection{Approximate kernel learning for person re-id}\label{subsec:sampling}

By projecting data points into high-dimensional or even infinite-dimensional feature space, kernel methods are found very effective in person re-id application with strong generalization performance \cite{Xiong2014Person}.
However, one limitation of kernel methods is their high computational cost, which is at least quadratic in the size of training samples, due to the calculation of kernel matrix. In the testing stage, the kernel similarity between any two identities cannot be computed directly but through their individual kernel similarities to training individuals. Thus, the time complexity of testing heavily depends on the number of training samples. For instance, in CUHK03 data set, the number of individuals in the training is 1260, which would cause inefficiency in testing. To avoid computing kernel matrix, a kernel learning problem is often approximated by a linear prediction problem where the kernel similarity between any two data points is approximated by their vector representations. Whilst both random Fourier features \cite{nips2007random} and the Nystr\"{o}m method \cite{Nystrom2005JMLR} have been successfully applied to efficient kernel learning, recent theoretical and empirical studies show that the Nystr\"{o}m method has a significantly better generalization performance than random Fourier features when there is a large eigengap of kernel matrix \cite{nips2012nystrom}.

The Nystr\"{o}m method approximates the kernel matrix by randomly selecting a subset of training examples and computes a kernel matrix $\widehat{K}$ for the random samples, leading to data dependent vector representations. The addditional error caused by the Nystr\"{o}m method in the generalization performance can be improved to be $O(1/m)$ ($m$ is the number of sampled training examples) when there is a large gap in the eigen-specturm of the kernel matrix \cite{nips2012nystrom}. Thus, we employ an improved Nystr\"{o}m framework to approximate the kernel learning in large-sized person re-id databases. Specifically, given a collection of $N$ training examples, $X=\{(\bx_1,y_1),\ldots,(\bx_N,y_N)\}$ where $x_i\in \mathcal{X} \subseteq\mathbb{R}^d$, $y_i\in \mathcal{Y}$. Let $\kappa(\cdot,\cdot)$ be a kernel function, $\mathcal{H}_k$ denote the endowed Reproducing Kernel Hilbert Space, and $K=[\kappa(\bx_i,\bx_j)]_{N\times N}$ be the kernel matrix for the samples in $X$. The Nystr\"{o}m method approximates the full kernel matrix $K$ by firstly sampling $m$ examples, denoted by $\hat{\bx}_1,\ldots,\hat{\bx}_m$, and then constructing a low rank matrix by $\widehat{K}_r$ where $r$ denotes the rank of $\widehat{K}=[\kappa(\hat{\bx}_i,\hat{\bx}_j)]_{m\times m}$. Thus, we can derive a vector representation of data by $\bz_n(\bx)=\widehat{D}_r^{-1/2}\widehat{V}_r^T(\kappa(\bx,\hat{\bx}_1),\ldots,\kappa(\bx,\hat{\bx}_m))^T$, where $\widehat{D}_r=diag(\hat{\lambda}_1,\ldots,\hat{\lambda}_r)$, $\widehat{V}_r=(\hat{\bv},\ldots,\hat{\bv}_r)$, $\hat{\lambda}_i$ and $\hat{\bv}_i$ denote the eigenvalue and eigenvector of $\widehat{K}_r$, respectively. Given $\bz_n(\bx)$, we aim to learn a linear machine $f(\bx)=\bw^T\bz_n(\bx)$ by solving the following optimization problem:
\begin{equation}
\min_{\bw\in\mathbb{R}^r}\frac{\lambda}{2}||\bw||_2^2 + \frac{1}{N}\sum_{i=1}^N l(\bw^T\bz_n(\bx_i),y_i).
\end{equation}

In \cite{nips2012nystrom}, Yang \etal construct an approximate functional space $\mathcal{H}_a=span(\hat{\varphi}_i,\ldots,\hat{\varphi}_r)$ where $\hat{\varphi}_i,\ldots,\hat{\varphi}_r$ are the first $r$ normalized eigenfunctions of the operator $L_m=\frac{1}{m}\sum_{i=1}^m\kappa(\hat{\bx}_i,\cdot)f(\hat{\bx}_i)$ and they subsequently obtain the following equivalent approximate kernel machine:
\begin{displaymath}
\min_{f\in\mathcal{H}_a}\frac{\lambda}{2}||f||^2_{\mathcal{H}_k}+\frac{1}{N}\sum_{i=1}^N l(f(\bx_i),y_i).
\end{displaymath}
According to both theoretical and empirical studies, the additional error caused by this approximation of the Nystr\"{o}m method is improved from $O(1/\sqrt{m})$ to $O(1/m)$ when there is a large gap between $\lambda_r$ and $\lambda_{r+1}$. In our experimental study (section \ref{ssec:kernel_app}), the eigenvalue distributions of kernel matrices from different person re-id data sets are shown to demonstrate the existence of large eigengap in kernel matrices.

\section{Experiments}\label{sec:exp}

\begin{figure}[t]
    \centering
        \includegraphics[width=2.5in]{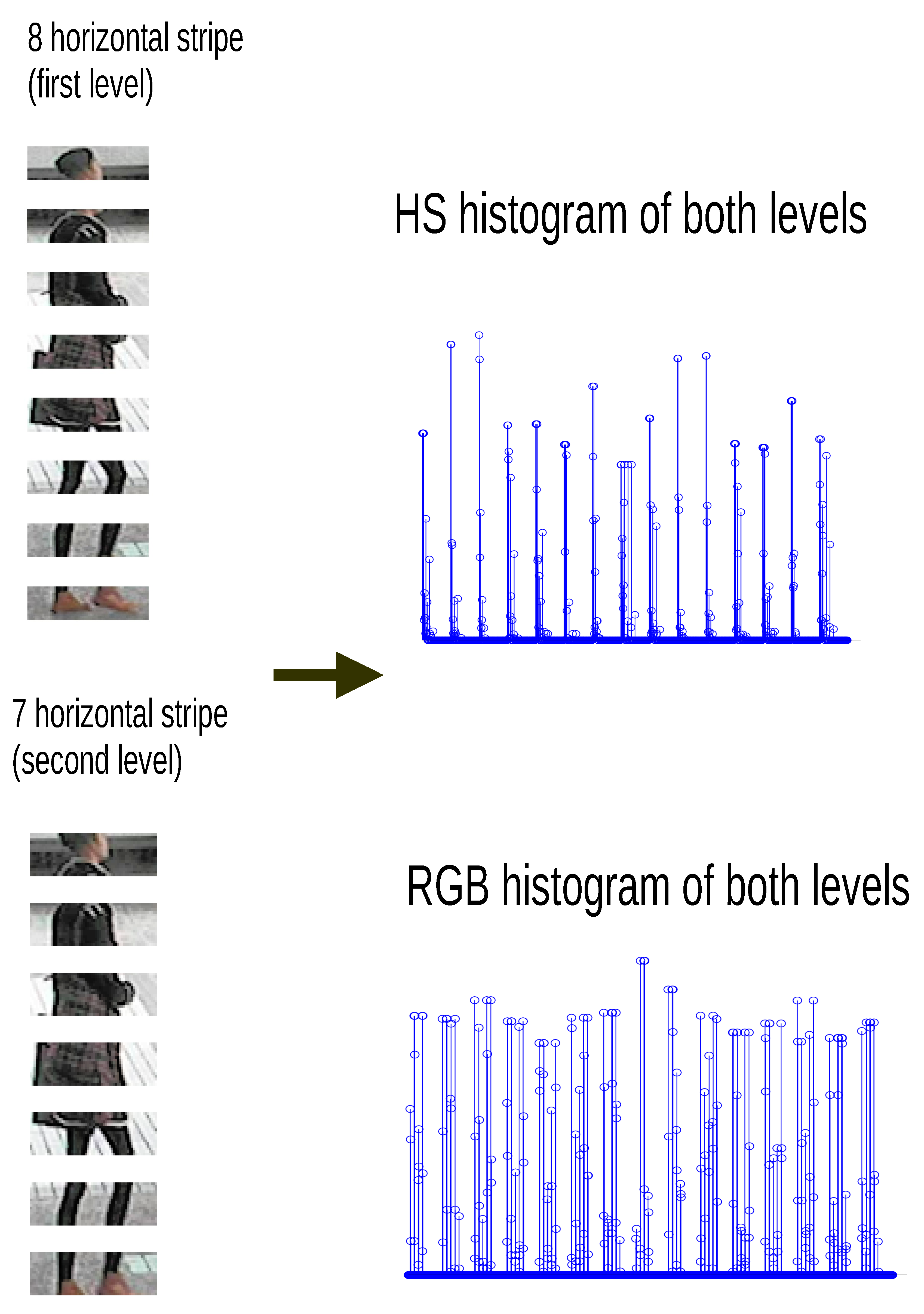}
    \caption{HS and RGB histograms extracted from overlapping stripes. }\label{fig:HS-RGB}
\end{figure}

\subsection{Visual feature implementations}\label{subsec:feat}

In this section, we introduce five low-level visual features which describe different aspects of person images, and can be well crafted to show promising results in person re-identification. To particularly cater for person re-id data sets, these feature descriptors are well manipulated in their implementations with configurations and settings best tuned.

\paragraph{LAB patterns}
For color patterns, we extract color features using $32$-bins histograms at three different scale factors ($0.5$, $0.75$ and $1$) over three different channels: L, A and B. Finally, each histogram feature is normalized with the $\ell_2$-norm and concatenated to form the final feature descriptor vector with length $4032$.

\paragraph{LBP patterns}
Local Binary Pattern (LBP) is an effective feature descriptor in describing local image texture and their occurrence histogram \cite{Ojala2002Multiresolution}. The standard $8$-neighbours LBP has a radius of $1$ and is formed by thresholding the $3 \times 3$ neighbourhood centred at a pixel's value. To improve the classification accuracy of LBP, we combine LBP histograms with color histograms extracted from the RGB colorspace.
The LBP$/$RGB implementation is obtained from \cite{Xiong2014Person}.
Specifically,  on each pedestrian image with resolution $48\times 128$ pixels, LBP and color histograms are extracted over a set of $7$ dense overlapping $48 \times 32$-pixels regions with a stepping stride of $16$ pixels in the vertical direction. For texture pattern, we extract LBP histograms using $8$-neighbours (radius $1$ and $2$) and
$16$-neighbours (radius $2$ and $3$). LBP is applied to grayscale image and each RGB color channel. We adopt an extension of LBP, known as the uniform LBP, which can better filter out noises \cite{Wang2009HOG}. For color histogram, we extract color features using $16$-bins histograms over six color channels: R, G, B, Y, Cb and Cr.
Finally, each histogram feature is normalized with the $\ell_1$-norm and concatenated to form a feature vector of $9352$ dimensions.

\paragraph{Hue-Saturation (HS) histogram}
We first re-scale an image to a broader size 64$\times$128, and then build a spatial pyramid by dividing the image into overlapping horizional stripes of 16 pixels in height. The HS histograms contain 8$\times$8 bins, which are computed for the 15 levels of the pyramids (i.e., 8 stripes for the first level plus 7 for the second level of overlapping stripes, where the second level of stripes are created from a sub-image by removing 8 pixels from top, bottom, left, and right of the original in order to remove background interference.). Each feature histogram is normalized with the $\ell_2$-norm and we have the vector length to be 960.

\paragraph{RGB histogram}
RGB is quantized into 8$\times$8$\times$8 over R, G, B channels. Like the extraction of HS, RGB histograms are also computed for the 15 levels of the pyramids, and each histogram feature is normalized with the $\ell_2$-norm, forming 7680-dim feature descriptor. We illustrate the procedure of extracting HS and RGB histograms in Fig. \ref{fig:HS-RGB}. Horizontal stripes capture information about vertical color distribution in the image, while overlapping stripes maintain color correlation between adjacent stripes in the final descriptor \cite{Giuseppe2015PAMI}. Overall, HS histogram renders a portion of the descriptor invariant to illumination variations, while the RGB histograms capture more discriminative color information, especially for dark and greyish colors.

\paragraph{SIFT features}
Scale-invariant feature transform (SIFT) is widely used for recognition task due to its invariance to scaling, orientation and illumination changes \cite{Lowe2004Distinctive}. The descriptor represents occurrences of gradient orientation
in each region. In this work, we employ discriminative SIFT  extracted from the LAB color space.
On a person image with a resolution of $48 \times 128$ pixels, SIFT  are extracted over a set of $14$ dense overlapping $32 \times 32$-pixels regions with a stepping stride of $16$ pixels in both directions. Specifically, we divide each region into $4 \times 4$ cells and set the number of orientation bins to $8$. As a result, we can obtain the feature vector with length 5376. In this paper, we use the SIFT$/$LAB implementation obtained from \cite{Zhao2013Unsupervised}.

\subsection{Experimental settings}
\paragraph{Datasets}
In this experiment,  six publicly available person re-identification data sets, iLIDS, 3DPES, PRID\-2011, VIPeR, CU\-HK\-01 and CU\-HK\-03 are used for evaluation. The iLIDS data set has $119$ individuals captured from eight cameras with different viewpoints \cite{Zheng2009Associating}.
The 3DPeS data set contains numerous video sequences taken from a real surveillance environment with eight different surveillance cameras and
consists of $192$ individuals \cite{Baltieri20113dpes}. The Person RE-ID 2011 (PRID2011) data set consists of images extracted from multiple person trajectories recorded from two surveillance
static cameras \cite{Hirzer2011Person}. Camera view A contains $385$ individuals, camera view B contains $749$ individuals,
with $200$ of them appearing in both views. Hence, there are $200$ person image pairs in the dataset.  VIPeR \cite{Gray2007Evaluating} contains $632$ individuals taken from two cameras with arbitrary viewpoints and varying illumination conditions.
The CUHK01 data set contains $971$ persons captured from two camera views in a campus environment \cite{Li2012Human} where one camera captures the frontal or back view of the individuals while another camera captures the profile view.
Finally, the CUHK03 data set consists of $1467$ persons taken from six cameras \cite{Li2014Deep}. The data set consists of manually cropped pedestrian images and images cropped from the pedestrian detector of \cite{Felzenszwalb2010Object}.
In our experiment, we use images which are manually annotated.

\begin{figure*}[t]
    \centering
        \includegraphics[width=0.3\textwidth,clip]{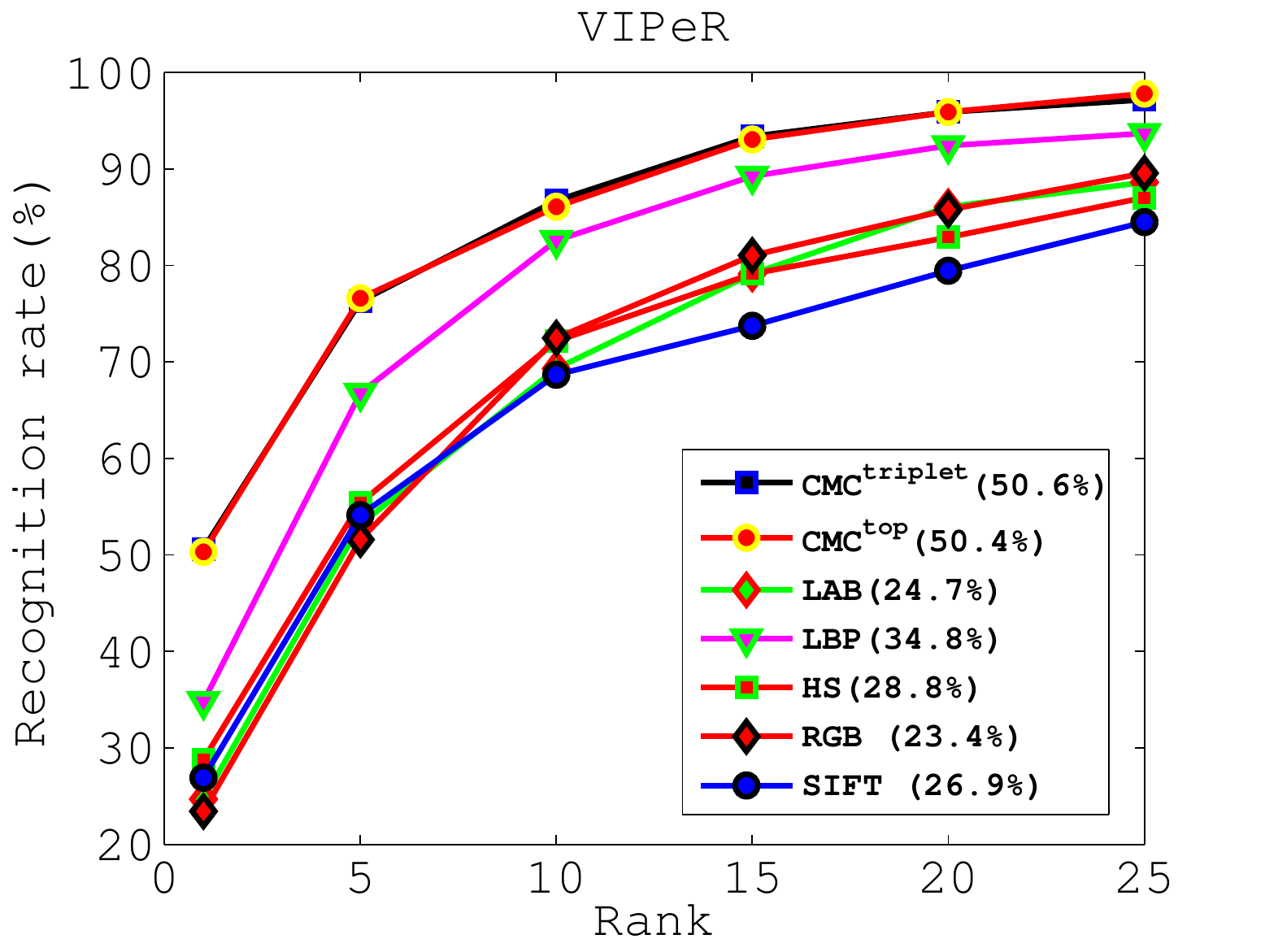}
        \includegraphics[width=0.3\textwidth,clip]{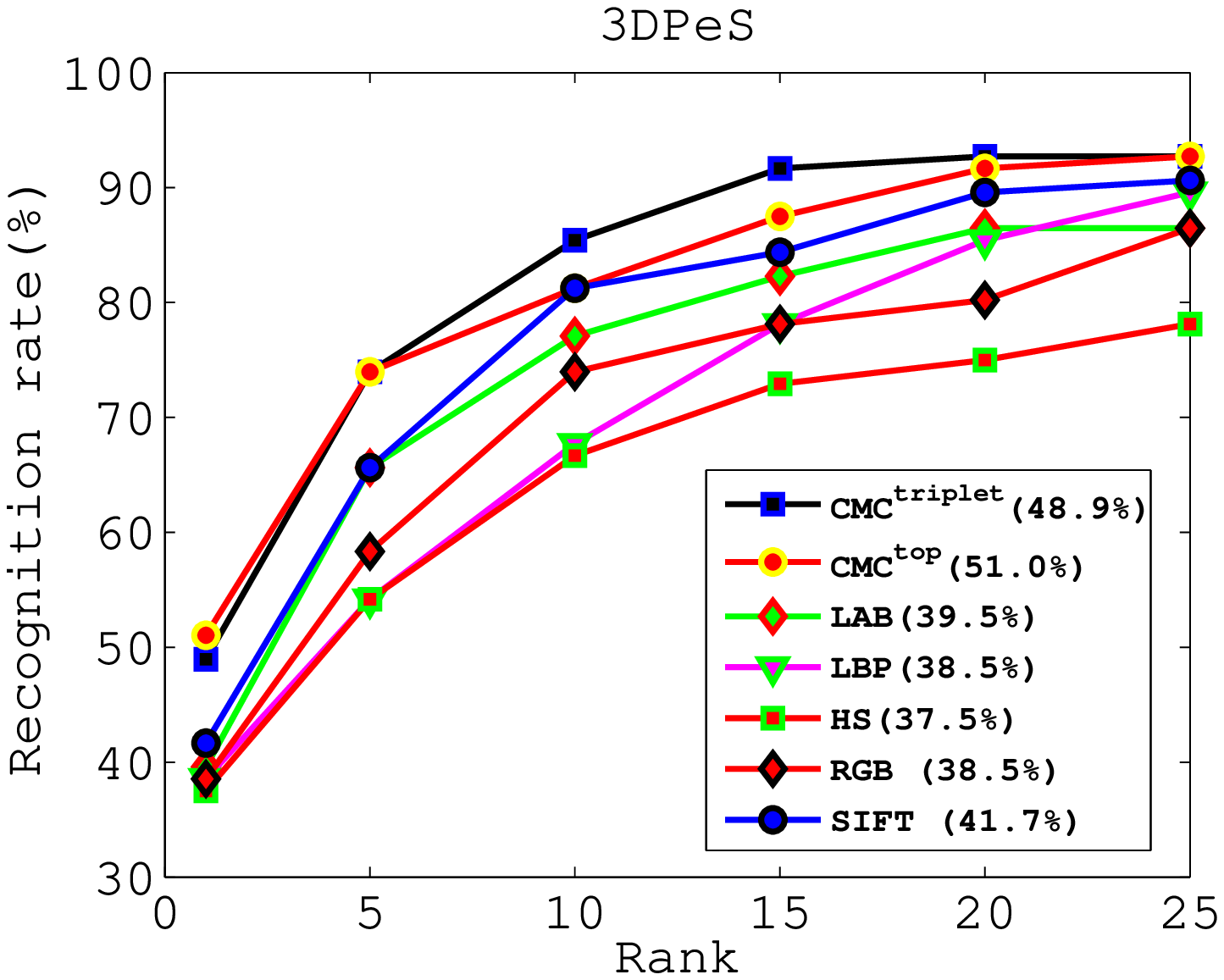}
        \includegraphics[width=0.3\textwidth,clip]{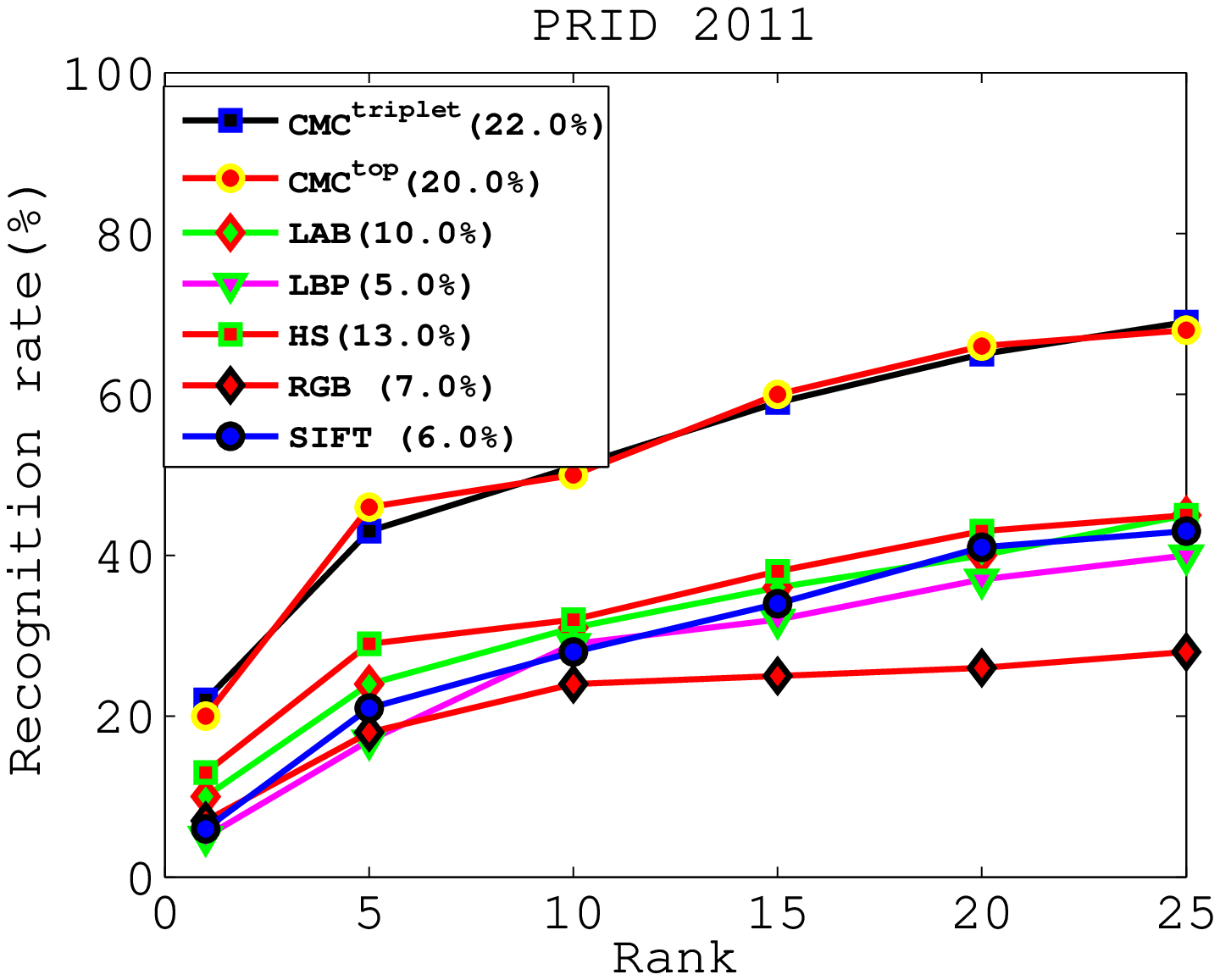}
        \includegraphics[width=0.3\textwidth,clip]{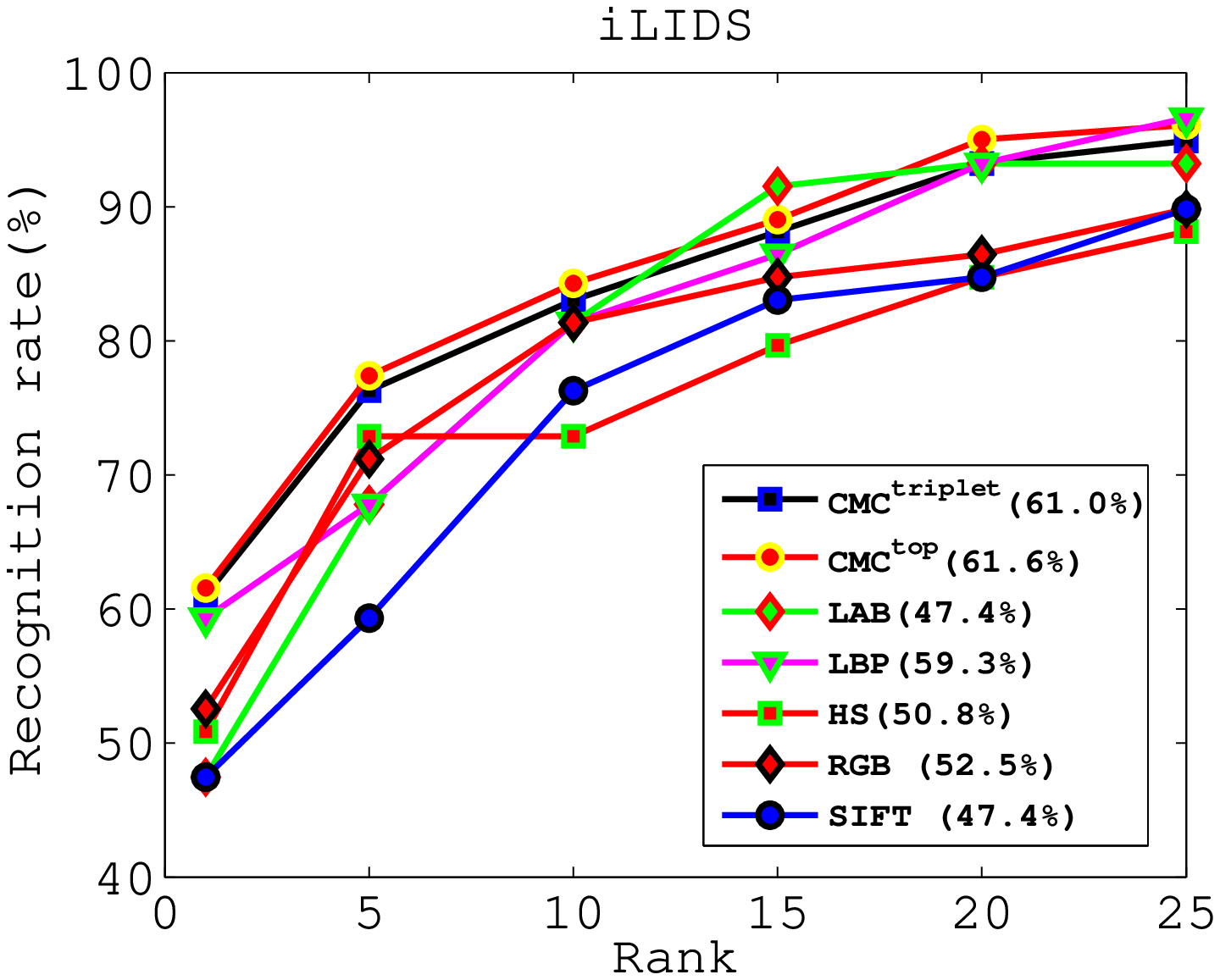}
        \includegraphics[width=0.3\textwidth,clip]{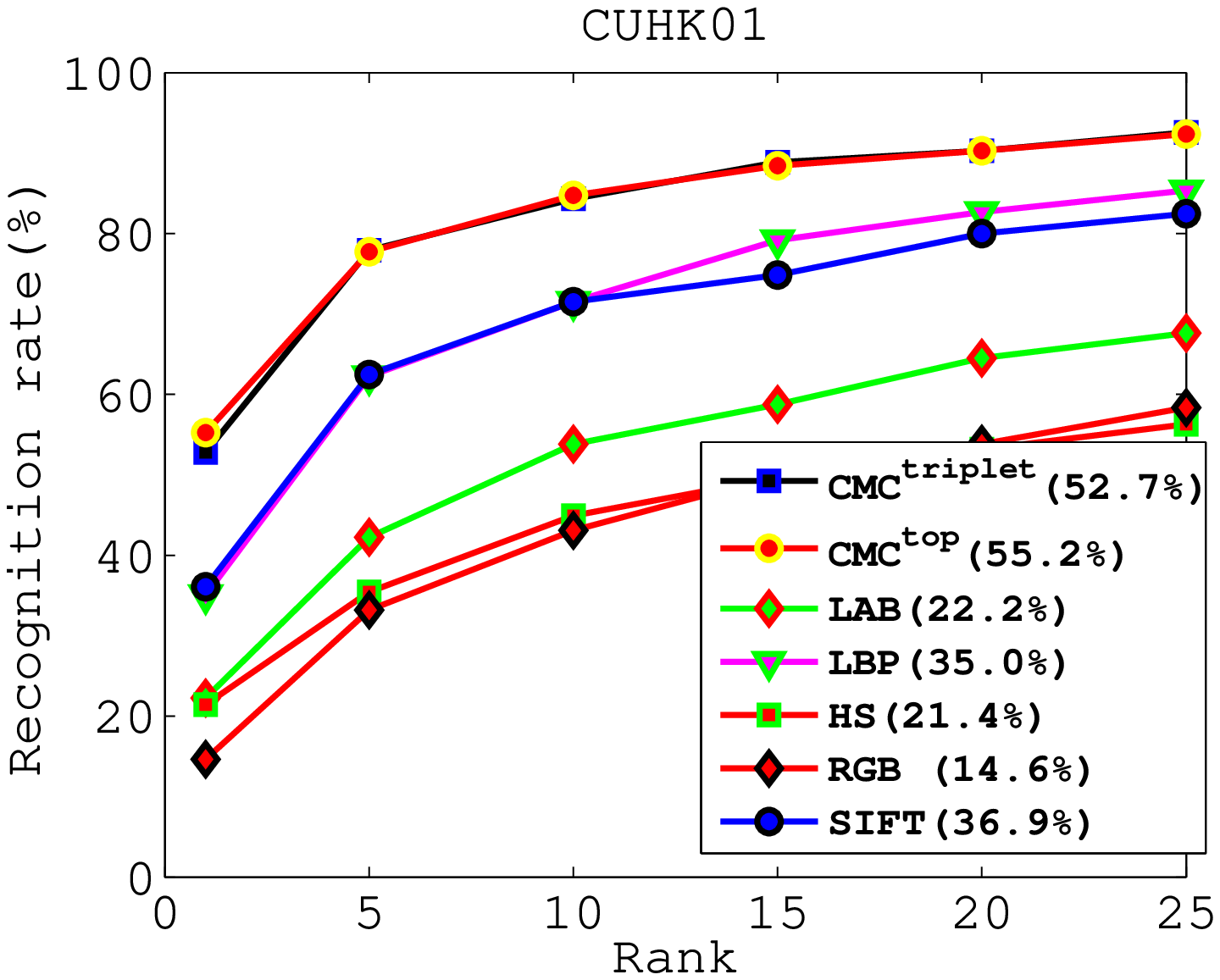}
        \includegraphics[width=0.3\textwidth,clip]{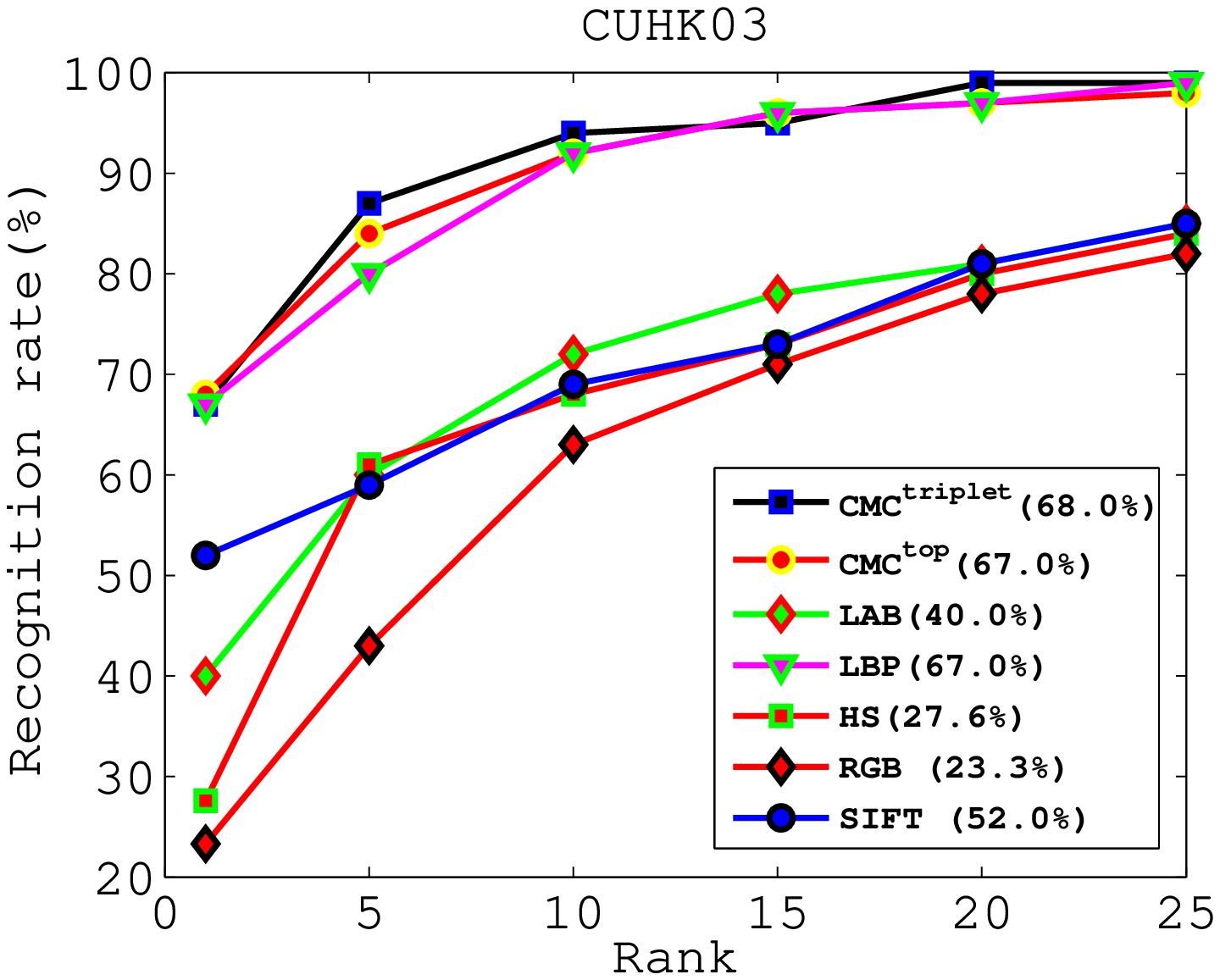}
    \caption{Performance comparison of base metrics (non-linear) with different visual features: LAB, LBP, HS, RBG histogram, and SIFT.  Rank-$1$ recognition rates are shown in parentheses.   The higher the recognition rate, the better the performance. \CMCstruct represents our ensemble approach which optimizes the CMC score over the top $k$ returned candidates. \CMCtriplet represents our ensemble approach which minimizes the number of returned candidates such that the rank-$k$ recognition rate is equal to one. } \label{fig:feat1}
\end{figure*}

\begin{figure*}[t]
    \centering
        \includegraphics[width=0.32\textwidth,clip]{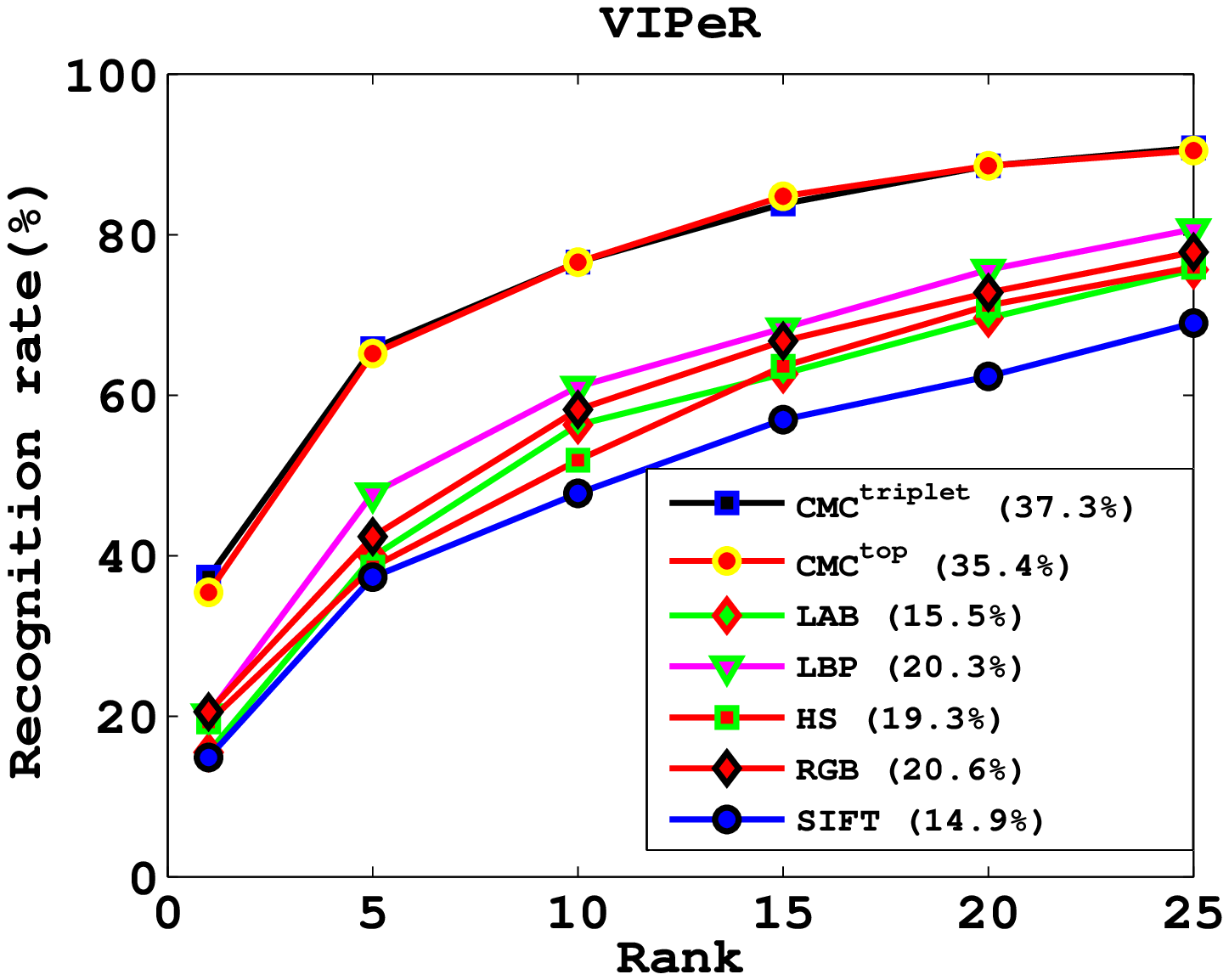}
        \includegraphics[width=0.32\textwidth,clip]{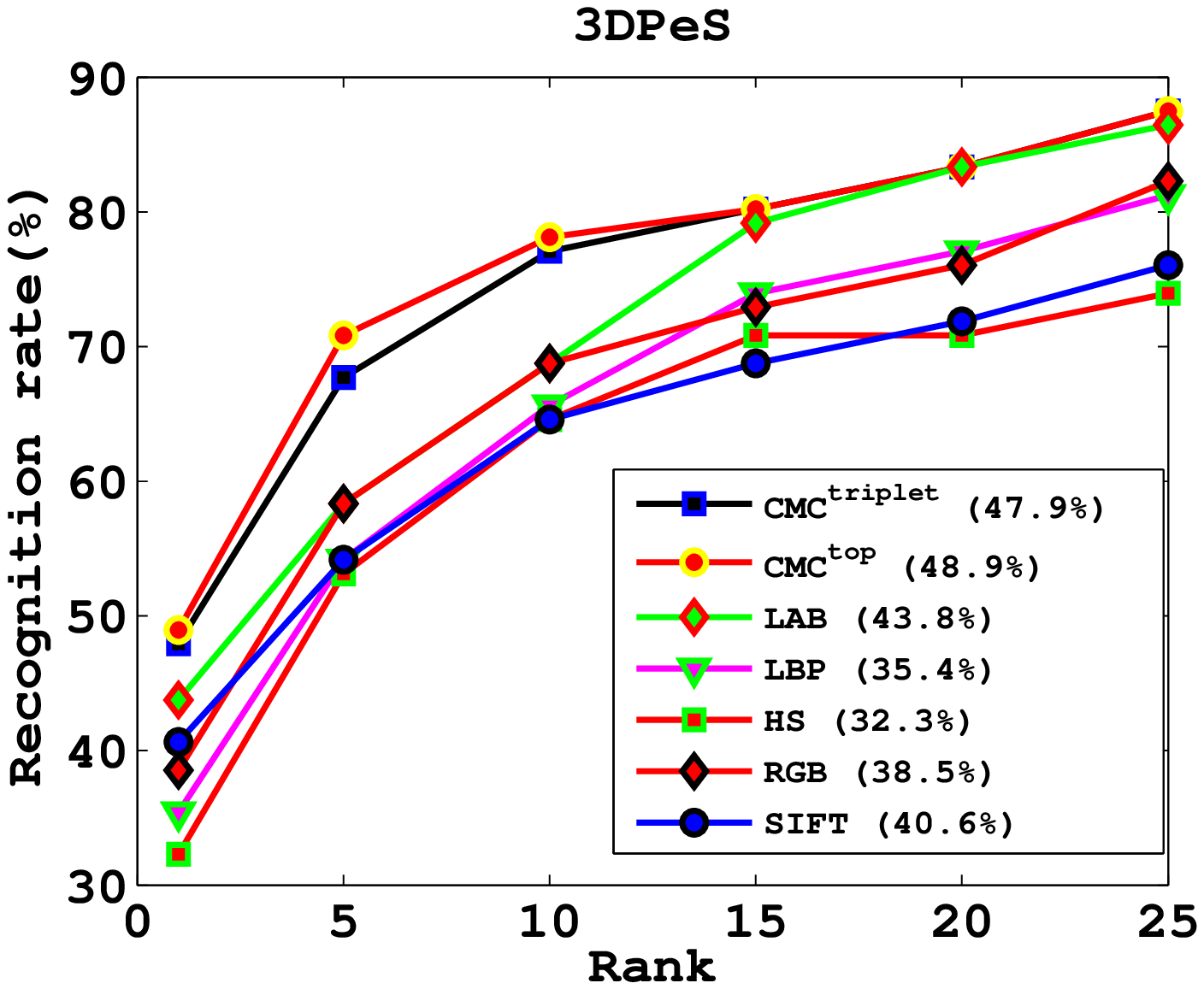}
        \includegraphics[width=0.32\textwidth,clip]{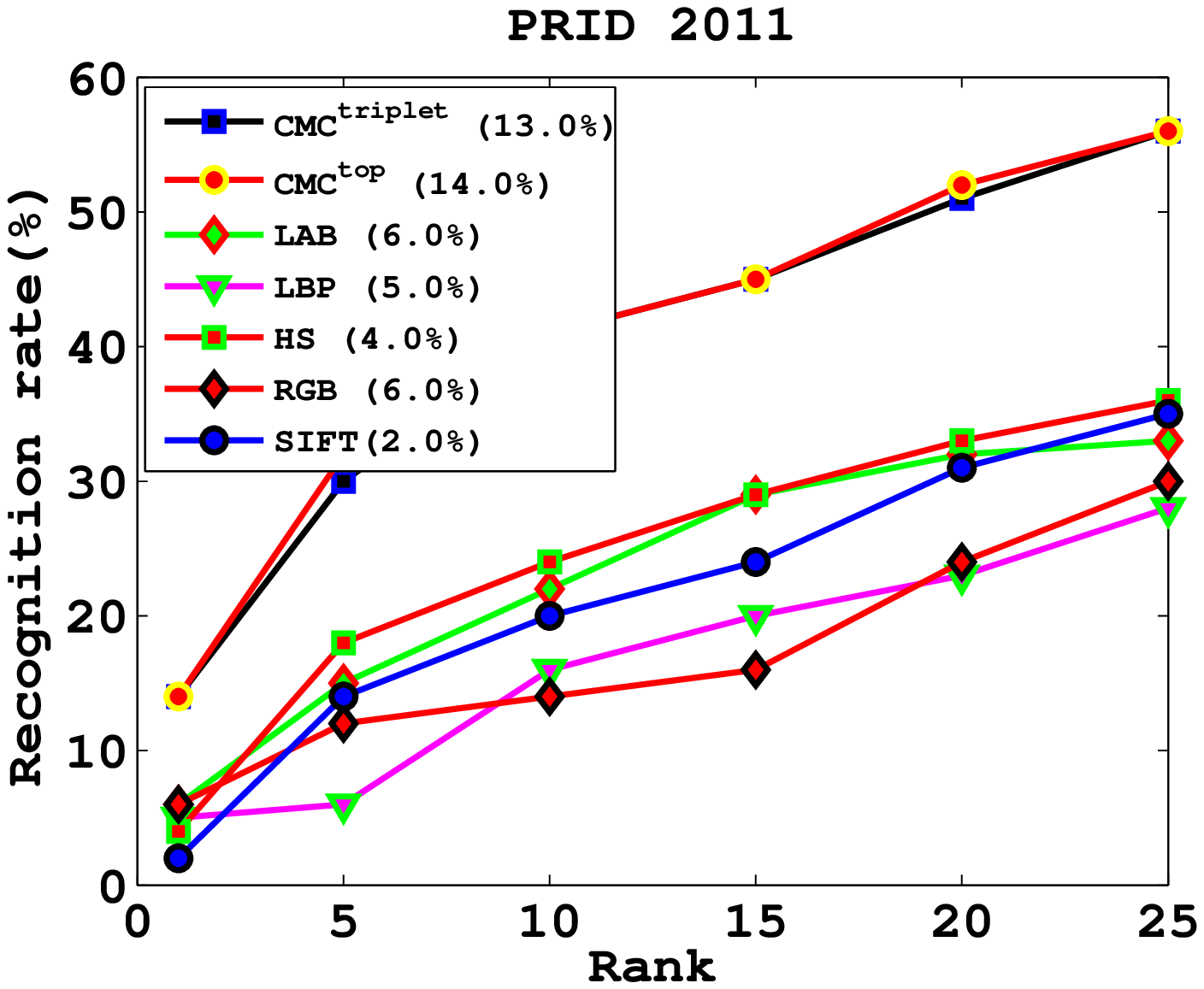}
        \includegraphics[width=0.32\textwidth,clip]{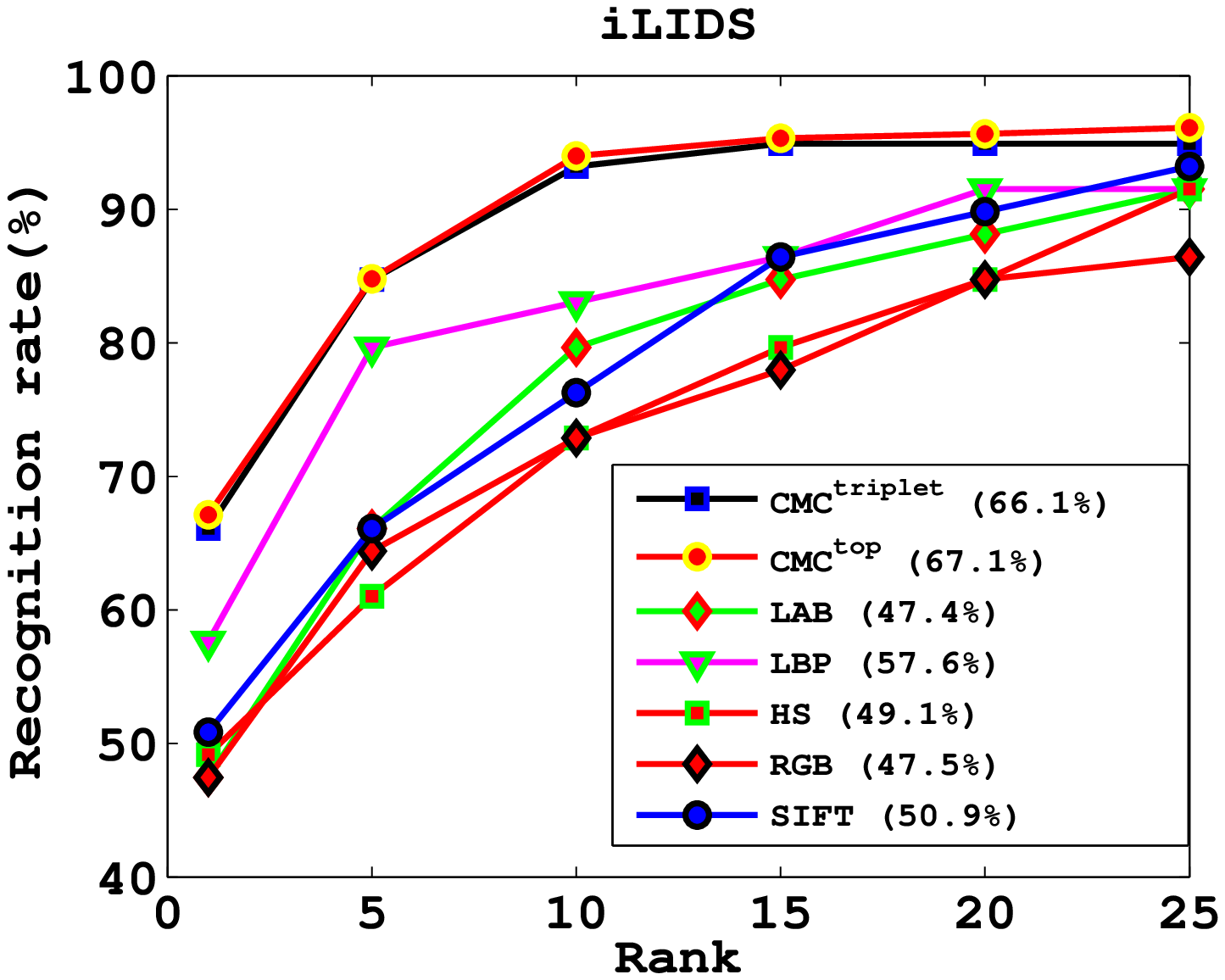}
        \includegraphics[width=0.32\textwidth,clip]{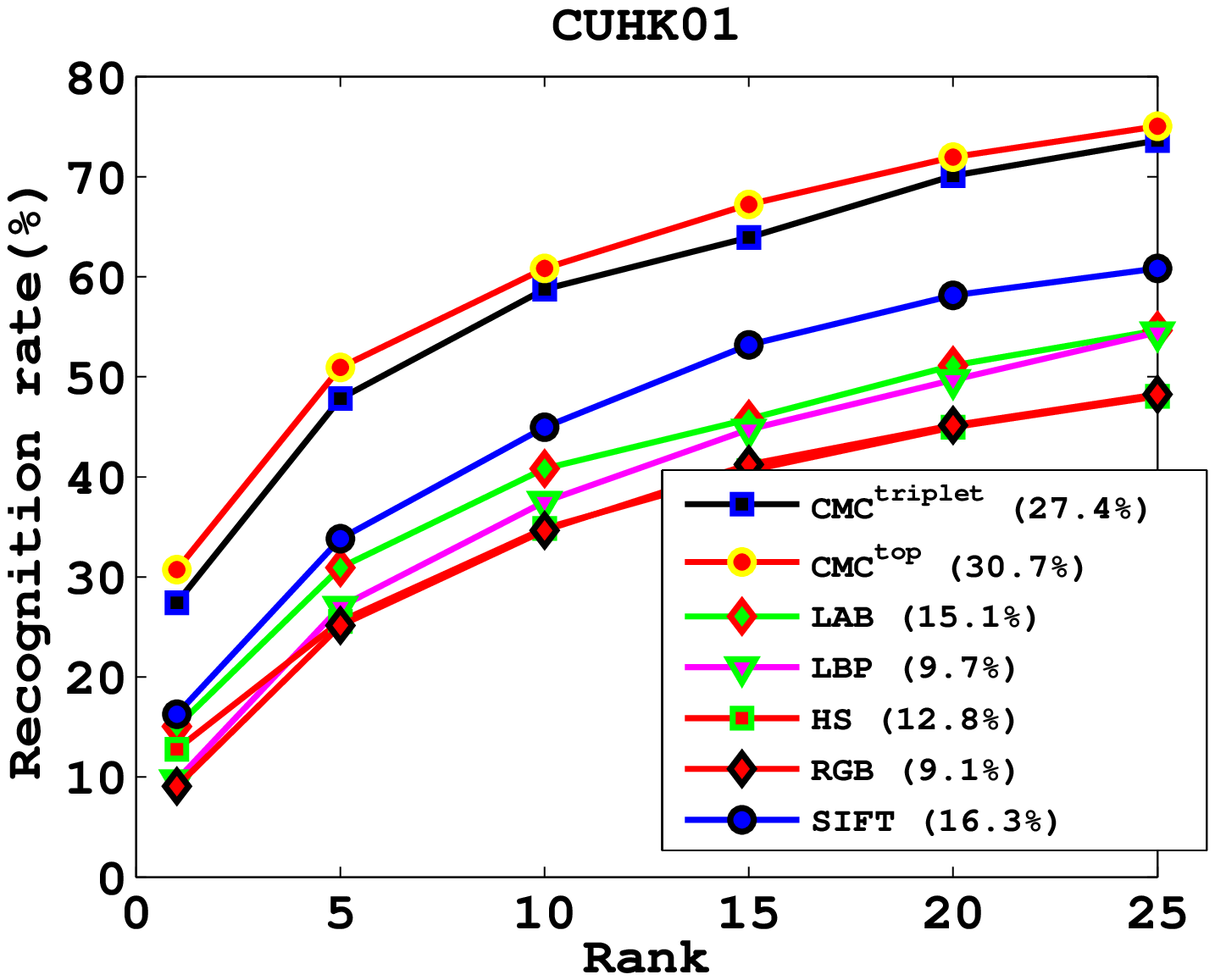}
        \includegraphics[width=0.32\textwidth,clip]{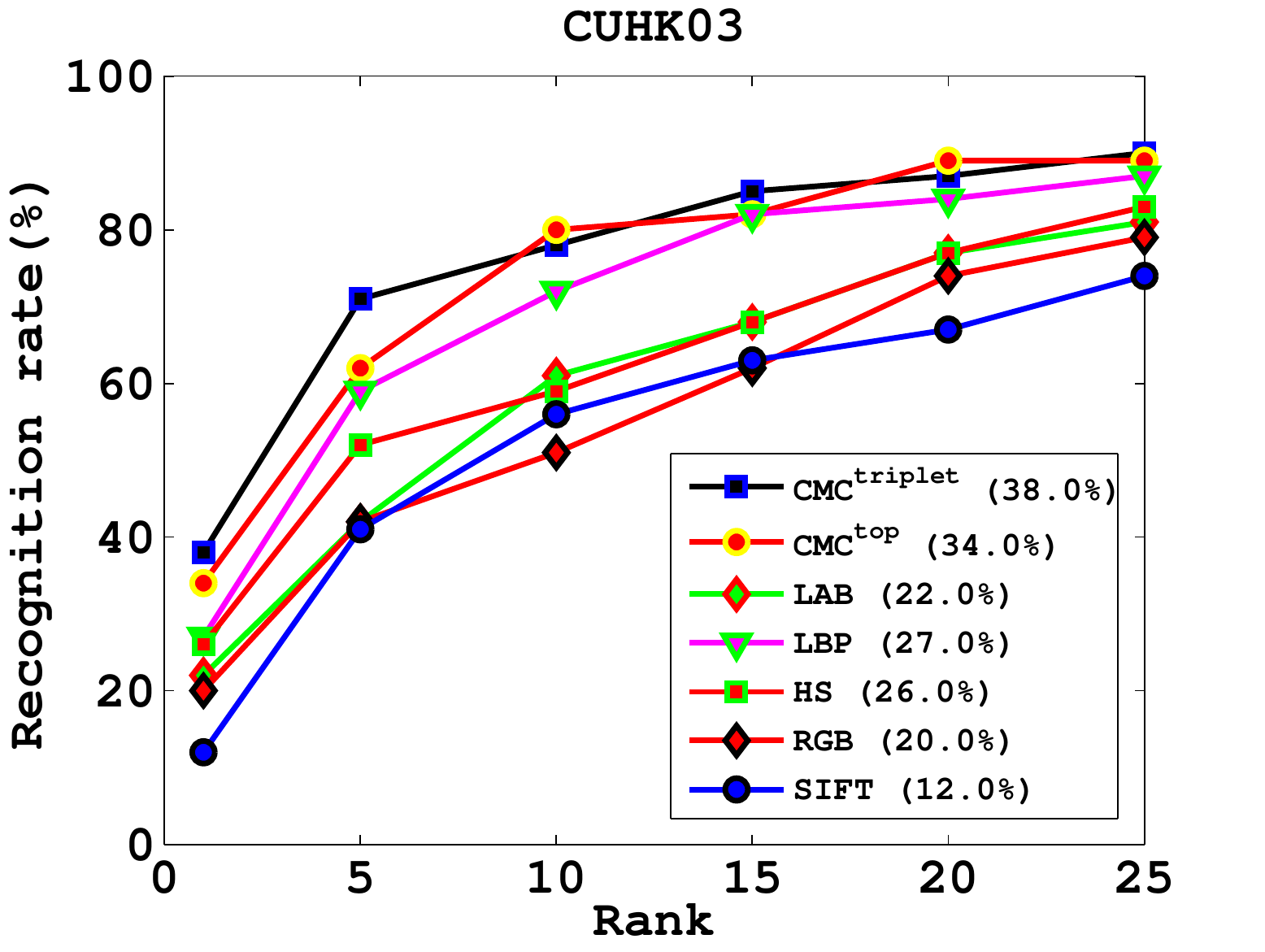}
    \caption{
    Performance comparison of base metrics (linear) with different visual features: LAB, LBP, HS, RBG histogram, and SIFT. Rank-$1$ recognition rates are shown in parentheses.  The higher the recognition rate, the better the performance.  \CMCstruct represents our ensemble approach which optimizes the CMC score over the top $k$ returned candidates.  \CMCtriplet represents our ensemble approach which  minimizes the number of returned candidates such that the rank-$k$ recognition rate is equal to one.}\label{fig:feat2}
\end{figure*}

\paragraph{Evaluation protocol}
In this paper, we adopt a single-shot experiment setting, similar to \cite{Li2013Learning,Pedagadi2013Local, Xiong2014Person, Zhao2014Learning, Zheng2011Person}. For all data sets except CUHK03, all individuals in the data set are randomly divided into two subsets so that training set and test set contains half of the individuals with no overlap on person identities.
For data set with two cameras, we randomly select two images of the same individual taken from two cameras as the probe image and the gallery image \footnote{In a multi-camera setup, the probe and gallery samples are defined as two samples which are randomly chosen from all the available cameras, one serving as the probe and the other as the gallery.}, respectively.
For multi-camera data sets, two images of the same individual are chosen: one is used as the probe image and the other as the gallery image. For CUHK03, we set the number of individuals in the train$/$test split to $1366$$/$$100$ as conducted in \cite{Li2014Deep}.
The split on training and test set on different data sets are shown in Table \ref{tab:best2}. This procedure is repeated $10$ times and the
average of cumulative matching characteristic (CMC) curves across $10$ partitions is reported.

\paragraph{Competitors} We consider the following state-of-the-art as competitors:
1) MidLevel \cite{Zhao2014Learning}; 2) LADF \cite{Li2013Learning}; 3) SalMatch \cite{Zhao2013SalMatch}; 4) KISSME \cite{Kostinger2012Large}; 5) PCCA \cite{PCCA}; 6) PRDC \cite{Zheng2013PAMI}; 7) eSDC \cite{Zhao2013Unsupervised}; 8) SDALF \cite{Farenzena2010Person}; 9) ISR \cite{Giuseppe2015PAMI}; 10) ITML \cite{Davis2007Information}; 11) LMNN \cite{Weinberger2006Distance}. Each algorithm is  tuned to show their best performance.

\paragraph{Parameters setting}
For the linear base metric (KISS ML \cite{Kostinger2012Large}), we apply principal component analysis (PCA) to reduce the dimensionality and remove noise.  For each visual feature, we reduce the feature dimension to $64$ dimensional subspaces. For the non-linear base metric (kLFDA \cite{Xiong2014Person}), we set the regularization parameter
for class scatter matrix to $0.01$, i.e., we add a small identity matrix to the class scatter matrix. For all features, we apply the RBF-$\chi^2$ kernel.  The kernel parameter is tuned to an appropriate value for each data set. In this experiment, we set the value of $\sigma^2$ to be the same as the first quantile of all distances \cite{Xiong2014Person}.
For \CMCtriplet, we choose the regularization parameter ($\nu$ in \eqref{EQ:svm}) from
$\{10^{3}$,$10^{3.1}$,$\cdots$,$10^{4}\}$ by cross-validation on the training data.
For \CMCstruct, we choose the regularization parameter ($\nu$ in \eqref{EQ:struct}) from $\{10^{2}$,$10^{2.1}$,$\cdots$,$10^{3}\}$ by cross-validation on the training data. We set the cutting-plane termination threshold to $10^{-6}$.
The recall parameter ($k$ in \eqref{EQ:featmap}) is set to be $10$ for iLIDS, 3DPeS, PRID2011 and VIPeR and $30$ for larger data sets (CUHK01 and CUHK03).

\subsection{Evaluation and analysis}

Since distance functions of different features have different scales, we normalize the distance between each probe image to all images in the gallery to be between zero and one. In other words, in all experiments, we set the distance between the probe image and the nearest gallery image to be zero and the distance between the probe image and the furthest gallery image to be one.

\paragraph{Feature evaluation}

We investigate the impact of low-level visual features on the recognition performance of person re-id.
Fig.~\ref{fig:feat1} shows the CMC performance of different visual features and their rank-$1$ recognition rates
when trained with the kernel-based LFDA on six benchmark data sets. We have the following observations:
\begin{itemize}
\item In VIPeR,  iLIDS, and CUHK03 data sets, LBP and LAB are more effective than other features. Since LBP encodes texture and color features, we suspect the use of color helps improve the overall recognition accuracy of LBP. In these data sets, a large number of persons wear similar types of clothing but with different color. Therefore, color information becomes an important cue for recognizing two different individuals. In some scenarios where color cue is not reliable because of illuminations, texture helps to identify the same individual.
\item In 3DPeS and CUHK01 data sets, SIFT and LBP are more effective in recognition. This can be ascribed to the fact that many samples are captured by zoomed/translated cameras (in 3DPeS) or orientation changes (in CUHK01) whilst SIFT is invariant to scale and orientation.
\item In PRID 2011 database, HS  becomes predominant feature. This is because this database consist of images captured with little exposure to illuminations in which both color and texture cues become unreliable whilst HS histograms are invariant to illumination changes.
\end{itemize}

\begin{figure*}[h!]
    \centering
        \includegraphics[width=0.3\textwidth,clip]{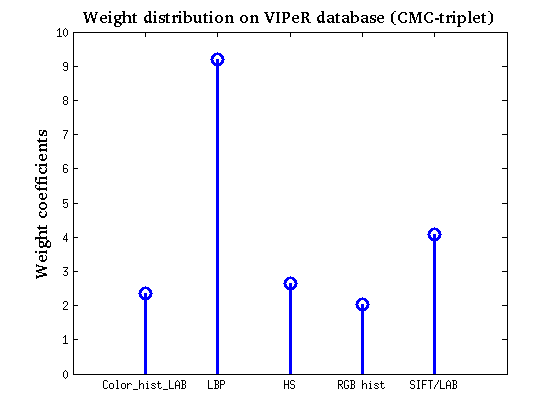}
        \includegraphics[width=0.3\textwidth,clip]{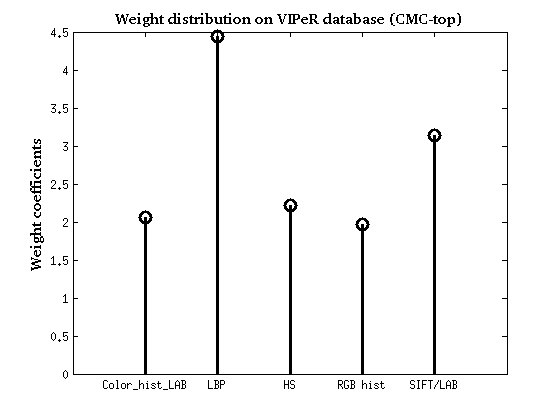}
        \includegraphics[width=0.3\textwidth,clip]{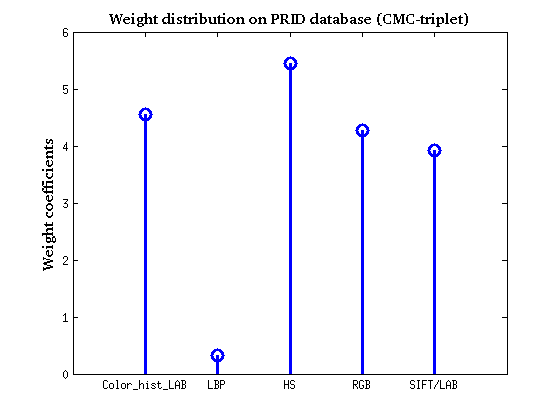}
        \includegraphics[width=0.3\textwidth,clip]{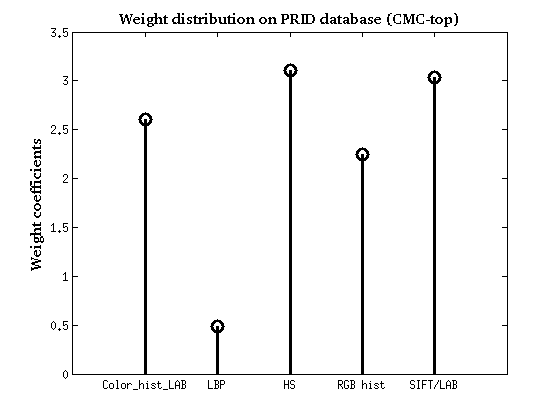}
        \includegraphics[width=0.3\textwidth,clip]{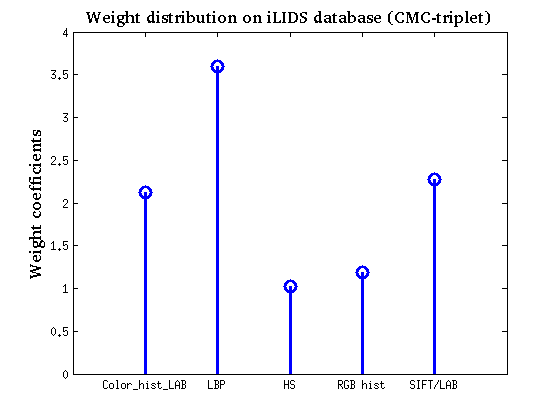}
        \includegraphics[width=0.3\textwidth,clip]{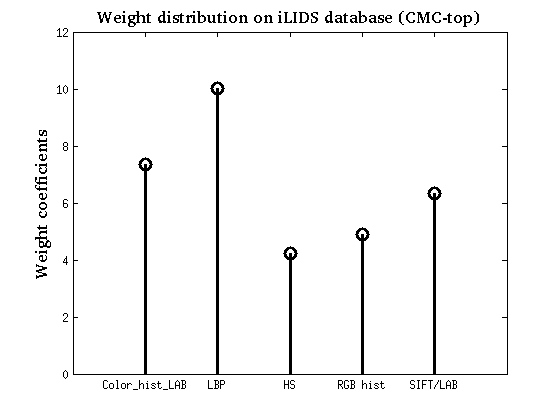}
        \includegraphics[width=0.3\textwidth,clip]{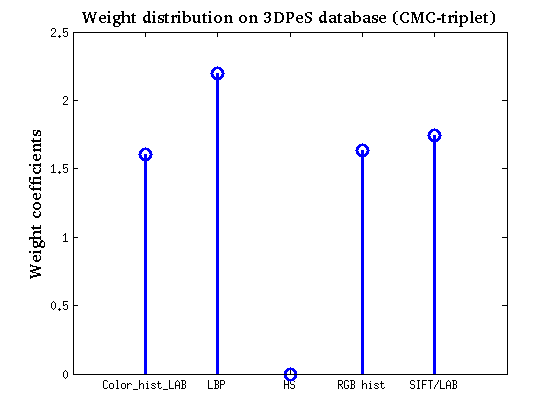}
        \includegraphics[width=0.3\textwidth,clip]{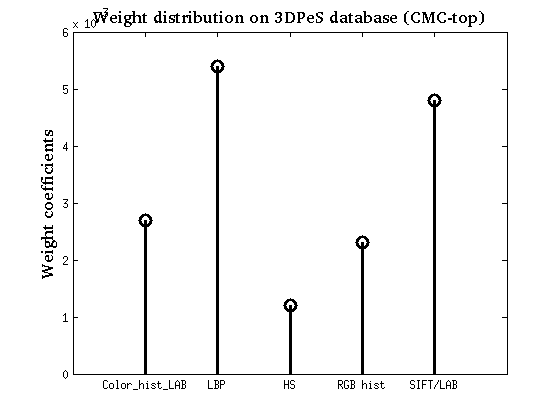}
        \includegraphics[width=0.3\textwidth,clip]{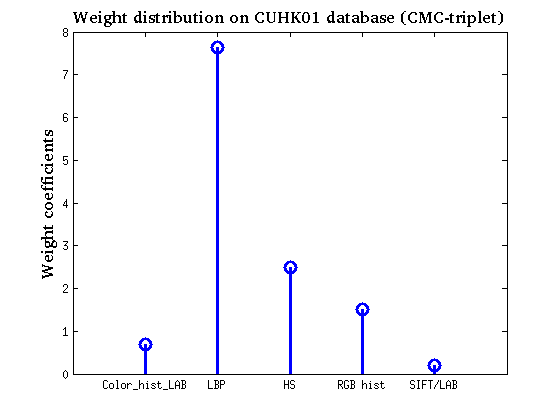}
        \includegraphics[width=0.3\textwidth,clip]{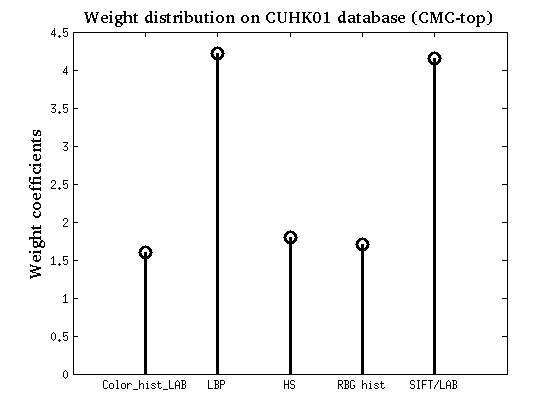}
        \includegraphics[width=0.3\textwidth,clip]{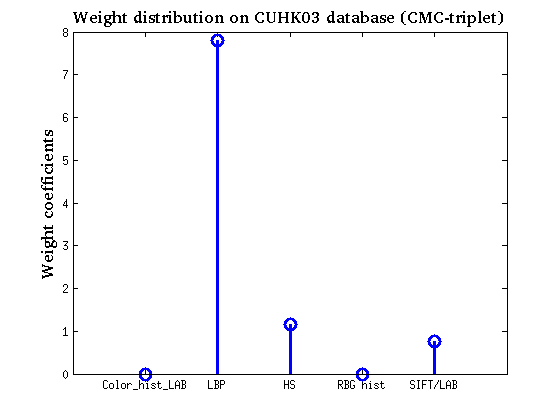}
        \includegraphics[width=0.3\textwidth,clip]{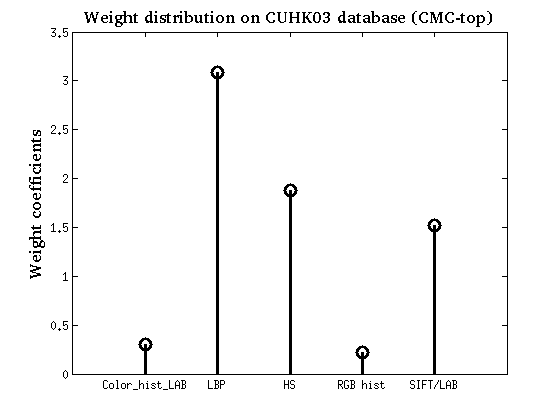}
    \caption{Weight distributions on six datasets that can reflect the adaptive effects of low-level features in recognizing individuals. The higher the coefficient is, the more important the feature is. It shows that the effectiveness of these features varied upon different datasets, which can be quantified in their weight values.} \label{fig:weight-distribute}
\end{figure*}

In fact, these person re-id data sets are created with distinct photometric and geometric transformations due to their own lighting conditions and camear settings. In this end, low-level features that are combined properly can be used to characterize the designed features of these data sets.  Since our proposed optimization yields a weight vector for distance functions, it is informative to characterize the nature of different test data sets and the effects of low-level features in recognizing persons are adaptive to these data sets. We show the weight distributions of corresponding features on six benchmarks in Fig. \ref{fig:weight-distribute}.  It can be seen that
\begin{itemize}
\item in VIPeR, iLIDS and 3DPeS data sets, the features of color, LBP and SIFT are more effective, thus they are more likely to be assigned larger weight coefficents.
\item In CUHK01 and CUHK03 databases, LBP, SIFT and HS are more helpful to recognize individuals due to the fact that many samples in the two data sets undergo photometric and geometric transforms caused by the change of lighting conditions, viewpoints, and human-pose whereas SIFT and HS are invariant to orientation and illumination changes, respectively.
\item In PRID 2011 data set, HS and color information (LAB and RGB) are predominant features.
\end{itemize}

We also provide  experimental results of {\em linear metric learning} \cite{Kostinger2012Large} with different visual features.
The CMC curve can be found in Fig.~\ref{fig:feat2}. Recall Fig. \ref{fig:feat1}, we observe that:
1) non-linear based kernels perform better than linear metrics on most visual features.
A significant performance improvement is found in CUHK03 data set, where we observe a performance improvement from $12.0\%$ to $52.0\%$ for SIFT features, and an improvement from $27.0\%$ to $67.0\%$ for LBP features;
2) our method is highly flexible to both linear/non-linear metric learning, and the effects of visual features in recognizing persons are almost the same in using linear or non-linear metric learning.

\begin{figure*}[t]
    \centering
        \includegraphics[width=0.3\textwidth,clip]{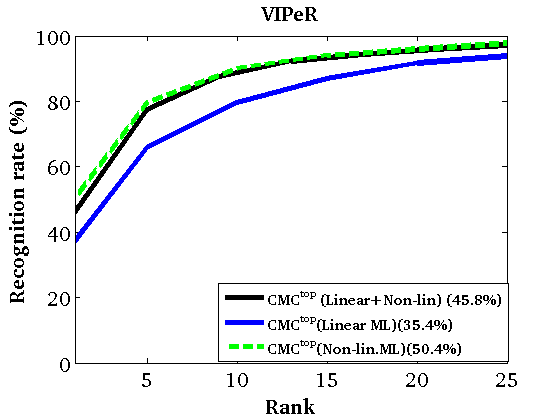}
        \includegraphics[width=0.3\textwidth,clip]{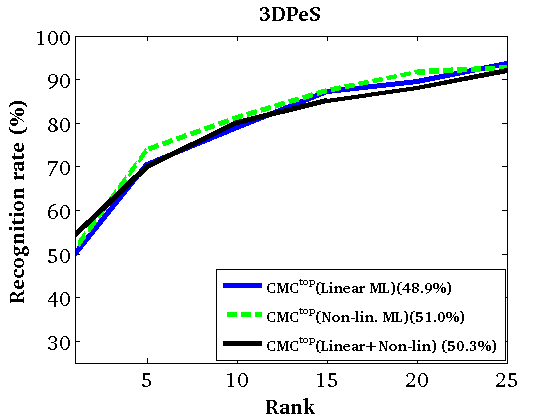}
        \includegraphics[width=0.3\textwidth,clip]{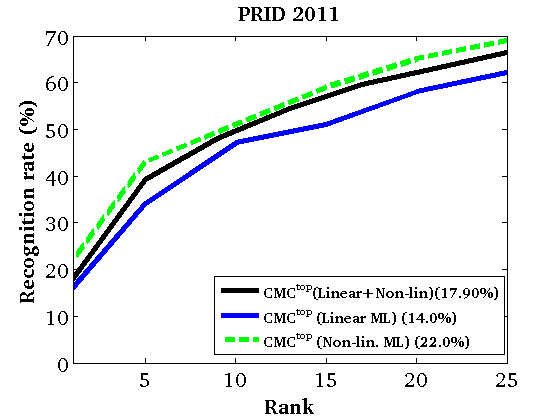}
        \includegraphics[width=0.3\textwidth,clip]{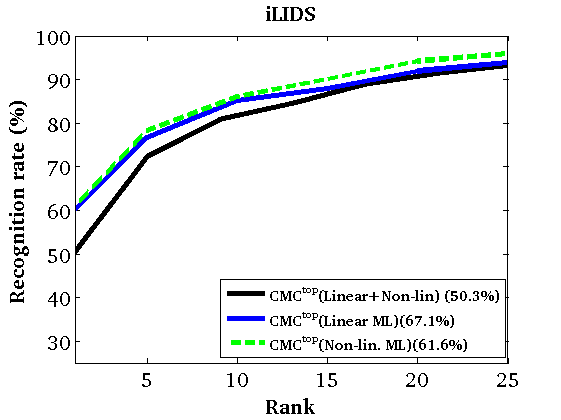}
        \includegraphics[width=0.3\textwidth,clip]{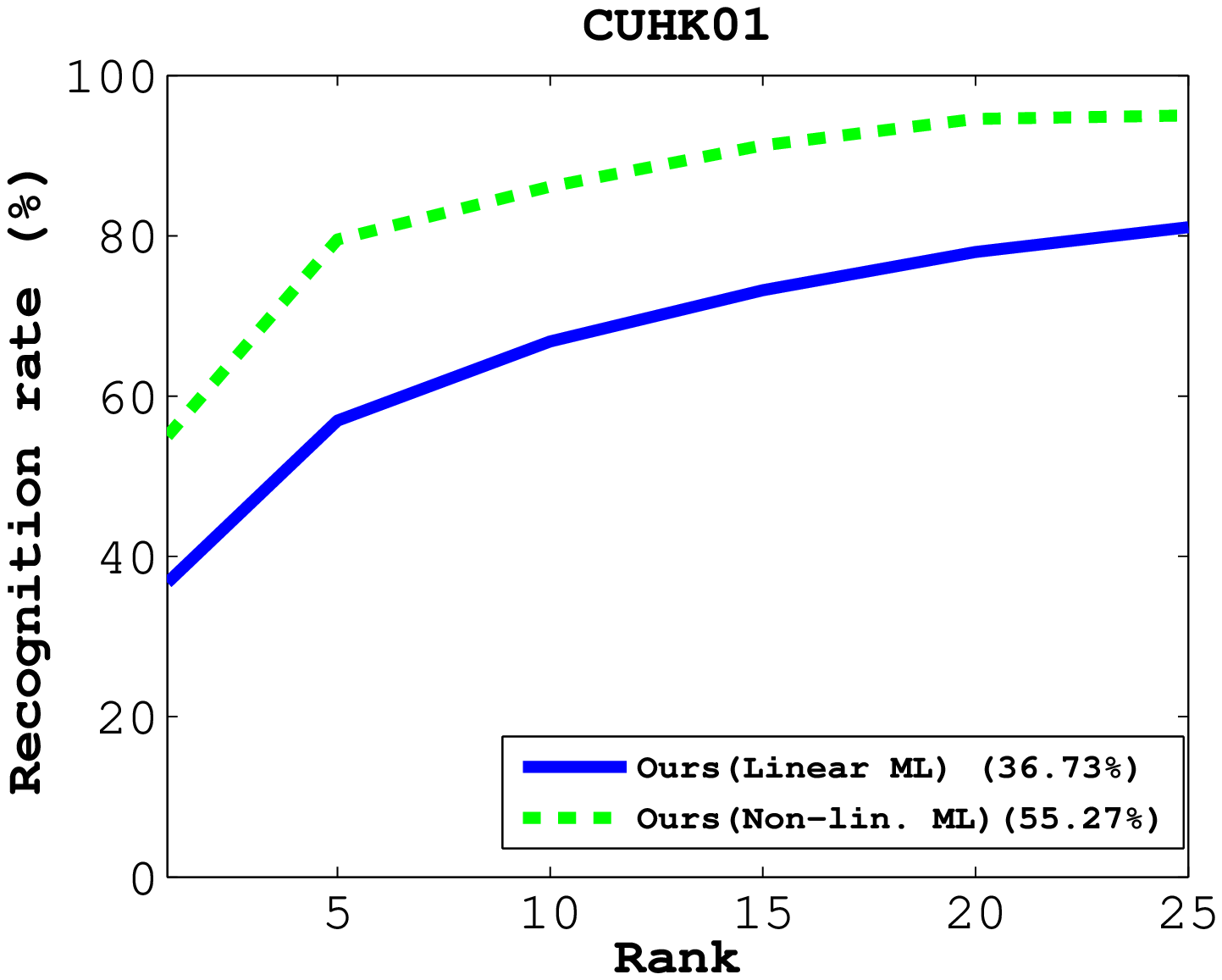}
        \includegraphics[width=0.3\textwidth,clip]{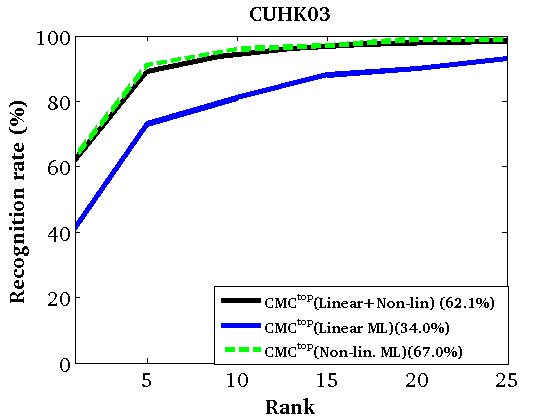}
    \caption{Performance comparison of \CMCstruct with two different base metrics: linear metric learning (KISSME) \cite{Kostinger2012Large} non-linear metric learning \cite{Xiong2014Person} (kLFDA with RBF-$\chi^2$ kernel), and their combination. On VIPeR, PRID 2011, CUHK01 and CUHK03 data sets,  an ensemble of non-linear base metrics significantly outperforms an ensemble of linear base metrics. }\label{fig:linear_nonlin}
\end{figure*}

\paragraph{Ensemble approach with linear/non-linear metric learning}
We conduct the performance comparison of our approach  with two different base metrics: linear metric learning (KISSME) \cite{Kostinger2012Large} and kernel metric learning (kLFDA) \cite{Xiong2014Person}.
In this experiment, we use \CMCstruct to learn an ensemble. Experimental results are shown in Fig.~\ref{fig:linear_nonlin}.
Some observations can be seen from the figure: 1) both approaches perform similarly when the number of train/test individuals is small, e.g., on iLIDS and 3DPeS; 2) non-linear base metric outperforms linear base metric by a large margin when the number of individuals increase as in CUHK01 and CUHK03 data sets. This is because  KISSME employs PCA to reduce the dimensionality of data since it requires solving a generalized eigenvalue problem of very large scatter $d\times d$ matrices where $d$ denotes the dimensionality of features. However, this dimensionality step, when applied to relatively diverse data set, can result in an undesirable compression of the most discriminative features. In contrast, kLFDA can avoid performing such decomposition, and flexible in choosing the kernel to improve the classification accuracy; 3) the combination of the two types of metrics cannot help improving the recognition rate.

\begin{table*}[tb]
  \centering
  \caption{Re-id recognition rate ($\%$) at different recall (rank). The best result is shown in boldface.  Both \CMCstruct and \CMCtriplet achieve similar performance when retrieving $\geq 50$ candidates.}\label{tab:cmc}
  {
  \begin{tabular}{l||c|c|c||c|c|c||c|c|c}
  \hline
  \multirow{2}{*}{Rank} & \multicolumn{3}{c||}{VIPeR} &
               \multicolumn{3}{c||}{CUHK01} &  \multicolumn{3}{|c}{CUHK03} \\
  \cline{2-10}
            & Avg. & \CMCtriplet & \CMCstruct &
                Avg. & \CMCtriplet & \CMCstruct & Avg. & \CMCtriplet & \CMCstruct   \\
  \hline
  \hline
$1$ & $50.4$ & $50.3$ & $\mathbf{50.6}$ & $52.5$ & $53.0$ & $\mathbf{55.2}$ & $59.5$   & $67.0$   & $\mathbf{68.1}$ \\
$2$ & $61.0$ & 61.4 & $\mathbf{61.4}$ & $63.5$ & $64.1$ & $\mathbf{64.9}$ & $71.9$   & $74.0$   & $\mathbf{77.8}$ \\
$5$ & $76.5$ & $77.3$ & $\mathbf{77.6}$ & $77.1$ & $76.7$ & $\mathbf{77.5}$ & $86.4$   & $85.8$   & $\mathbf{89.3}$ \\
$10$ & $88.4$ & $88.6$ & $\mathbf{88.9}$ & $84.3$ & $ \mathbf{84.8}$ & $84.5$ & $92.3$   & $92.5$   & $\mathbf{95.3}$ \\
$20$ & $95.9$ & 95.9 & $\mathbf{95.9}$ & $92.1$ & $91.7$ & $\mathbf{92.5}$ & $96.9$   & $\mathbf{99.0}$   & 97.0 \\
$50$ & $99.4$ & $\mathbf{99.7}$ & $99.5$ & $97.0$ & $\mathbf{97.1}$ & $96.9$ & $99.8$   & $99.7$  & $\mathbf{100.0}$ \\
$100$ & $99.9$ & 100.0 & $\mathbf{100.0}$ & $98.6$ & 98.6 & $\mathbf{98.6}$ & 100.0  & 100.0  & $\mathbf{100.0}$ \\
  \hline
  \end{tabular}
  }
\end{table*}

\begin{figure*}[t]
  \centering
        \includegraphics[width=0.43\textwidth,clip]{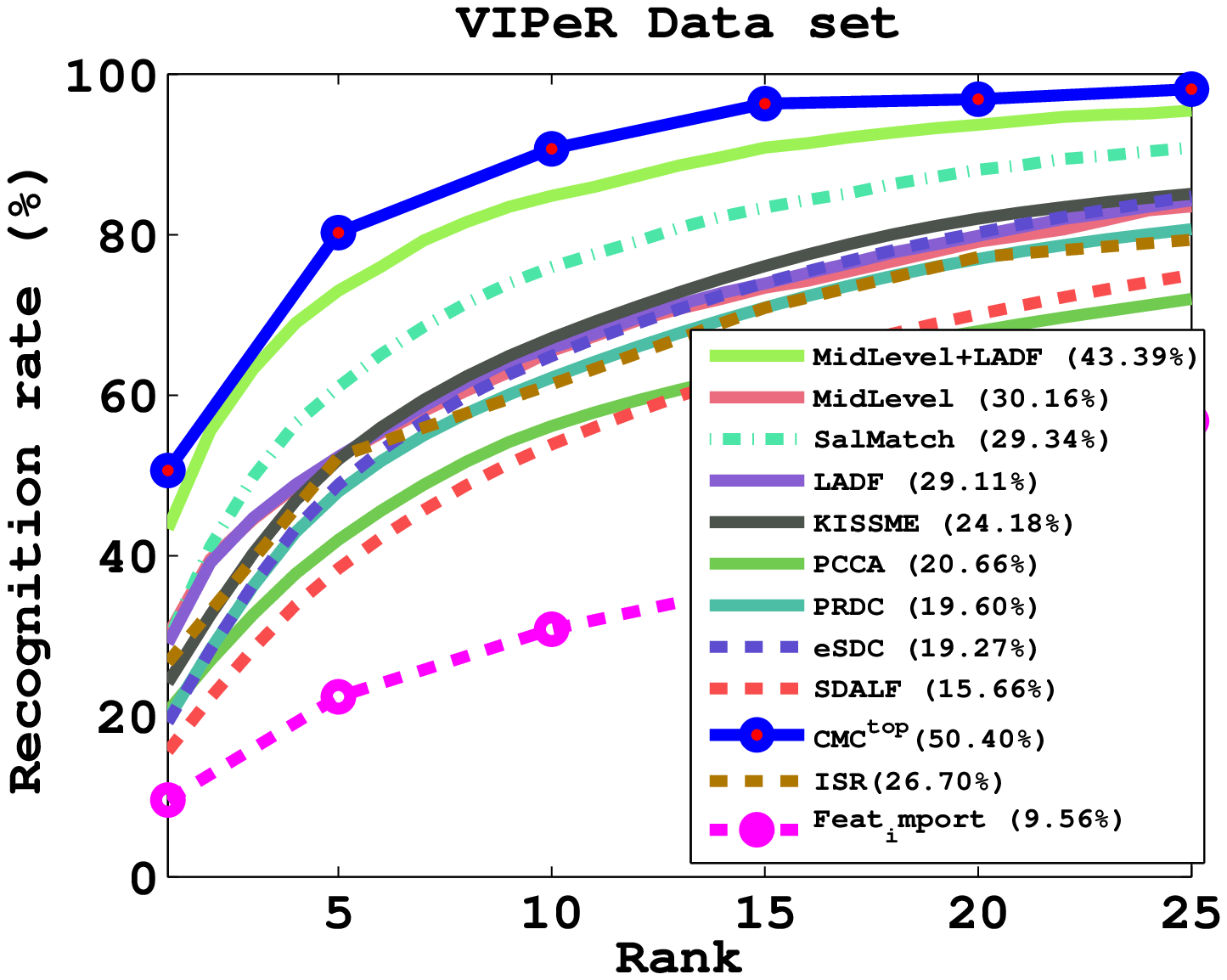}
        \includegraphics[width=0.43\textwidth,clip]{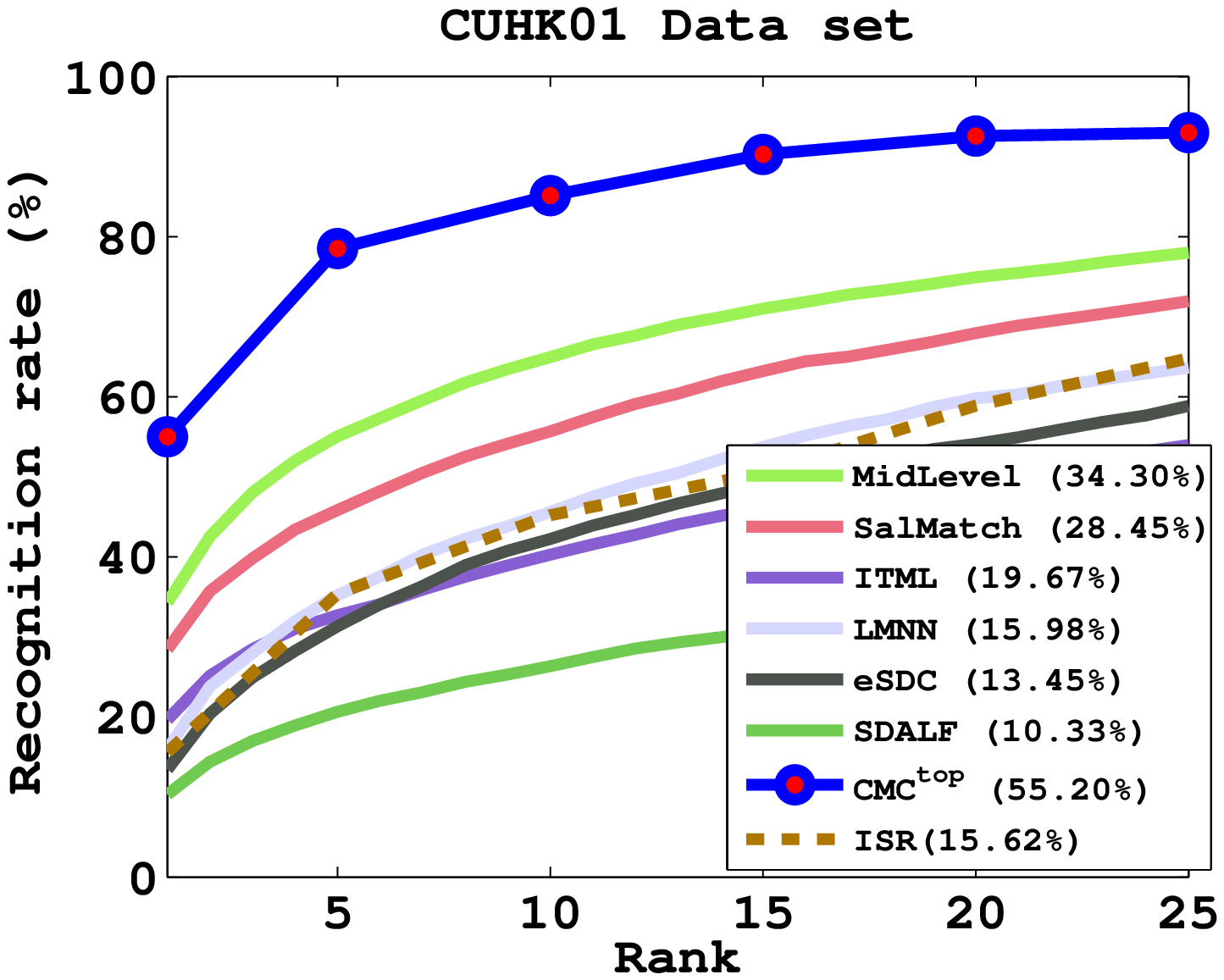}
    \caption{CMC performance for VIPeR and CUHK01 data sets. \CMCstruct outperforms all existing person re-id algorithms.} \label{fig:best1}
\end{figure*}

\begin{table}[t]
  \centering
\caption{Rank-$1$ recognition rate of existing best reported results and \CMCstruct results. }  \label{tab:best2}
  {
  \begin{tabular}{l|c|c|c|c}
  \hline
  \multirow{2}{*}{Data set} & \multicolumn{2}{|c|}{\# Individuals} &
                    \multirow{2}{*}{Prev. best}  & \multirow{2}{*}{\CMCstruct}\\
  \cline{2-3}
            & train & test &  &  \\
  \hline
  \hline
   iLIDS    &  $59$  & $60$  & $40.3\%$ \cite{Xiong2014Person}   & $\mathbf{61.6\%}$\\
   3DPeS    &  $96$  & $96$  & $\mathbf{54.2\%}$ \cite{Xiong2014Person}  & 51.1\%\\
   PRID2011 & $100$  & $100$ & $16.0\%$ \cite{Roth2014Mahalanobis}   & $\mathbf{22.0\%}$\\
   VIPeR    & $316$  & $316$ & $43.4\%$ \cite{Zhao2014Learning}   & $\mathbf{50.6\%}$\\
   CUHK01   & $486$  & $485$ & $34.3\%$ \cite{Zhao2014Learning}  & $\mathbf{55.2\%}$\\
   CUHK03   & $1366$ & $100$ & $20.7\%$ \cite{Li2014Deep}  & $\mathbf{68.0\%}$\\
  \hline
  \end{tabular}
  }
\end{table}

\paragraph{Performance at different recall values}
Next we compare the performance of \CMCtriplet with \CMCstruct. Both algorithms optimize the recognition rate of person re-id but with different objective criteria. kFLDA is employed as the base metric. The matching accuracy is shown in Table~\ref{tab:cmc}.
We observe that \CMCstruct achieves the best recognition rate performance at a small recall value. At a large recall value (rank $\geq 50$), both \CMCstruct and \CMCtriplet perform similarly. We also evaluate to see how \CMCstruct varies against different values of $k$ (optimize the rank-$k$ recognition rate). Results are shown in Table \ref{tab:k}.

\begin{table}[h]\scriptsize
  \caption{Rank-1 recognition rate of \CMCstruct with varied value of $k$. } \label{tab:k}
  {
  \begin{tabular}{l|c|c|c|c|c|c}
  \hline
    $k$  & VIPeR & PRID 2011 & 3DPeS& iLIDS & CUHK01 & CUHK03  \\
  \hline
  \hline
   10 & 50.4\% & 20.0\% & 51.0\% & 61.6\% & - & -\\
   20 & 50.4\% & 20.0\% & 51.2\% & 61.3\% & - & -\\
   30 & 49.8\% & 19.4\% & 50.7\% & 61.0\% & 55.2\% & 67.1\%\\
  40 & 48.4\% & 18.4\% & 50.1\% & 60.3\% & 55.2\% & 67.0\%\\
  50 & 48.1\% & 18.2\% & 48.4\% & 58.1\% & 55.0\% & 66.8\%\\
  \hline
  \end{tabular}
  }
\end{table}

\subsection{Comparison with state-of-the-art results}

Fig.~\ref{fig:best1} compares \CMCstruct (kLFDA is used as the base metric) with other person re-id algorithms on two major benchmark data sets: VIPeR and CUHK01.
Our approach outperforms all existing person re-id algorithms. Then we compare \CMCstruct with
the best reported results in the literature, as shown in Table~\ref{tab:best2}.
The algorithm proposed in \cite{Xiong2014Person} achieves state-of-the-art results on iLIDS and 3DPeS data sets ($40.3\%$ and $54.2\%$ recognition rate at rank-$1$, respectively). \CMCstruct outperforms  \cite{Xiong2014Person}  on the iLIDS ($61.3\%$) by a large margin, and achieves a comparable result on 3DPeS (51.1\%). Zhao \etal propose mid-level
filters for person re-identification \cite{Zhao2014Learning}, which achieve state-of-the-art results on the VIPeR and CUHK01 data sets
($43.39\%$ and $34.30\%$ recognition rate at rank-$1$, respectively). \CMCstruct outperforms \cite{Zhao2014Learning} by
achieving a recognition rate of $50.6\%$ and $55.2\%$ on the VIPeR and CUHK01 data sets, respectively.

\subsection{Comparison with a structured learning-based person re-id}
In this section, we compare our ensemble approaches with another structured learning-based method for
person re-\-id\-en\-ti\-fi\-ca\-tion \cite{Wu2011Optimizing}. In \cite{Wu2011Optimizing}
a metric learning to rank (MLR) algorithm \cite{McFee2010Metric} is applied to the person re-id\-en\-ti\-fi\-ca\-ti\-on problem with a listwise loss function and
evalued on Mean Reciprocal Rank (MRR). For a fair comparison, we concatenate low-level visual features and learn MLR \cite{Wu2011Optimizing} and LMNN \cite{Weinberger2006Distance} on VIPeR data set.
We choose the MLR trade-off parameter from $\{10^{-5},10^{-4},\cdots,10^{5}\}$.  The rank-$1$ recognition rate and mean reciprocal rank (MRR)  are shown in the Table~\ref{tab:ens}. We can see that our ensemble approach significantly outperforms both baseline methods under both evaluation criteria.

\begin{table}[bt]
  \centering
  \caption{Rank-$1$ recognition rate and mean reciprocal rank (MRR) of different algorithms. The experimental result reported here for MLR+MRR is better than  the one reported in \cite{Wu2011Optimizing} as the authors only use color histogram feature. }  \label{tab:ens}
  {
\begin{tabular}{l|c|c}
  \hline
    Algorithm  & Rank-$1$ rate & MRR \\
  \hline\hline
   LMNN \cite{Weinberger2006Distance}  & $24.9\%$ & $37.8\%$\\
   MLR+MRR \cite{Wu2011Optimizing}   &  $26.2\%$ &  $39.9\%$ \\
   \CMCstruct + linear metric  &  $37.3\%$ &  $47.3\%$ \\
   \CMCstruct + non-linear metric  &  $50.6\%$ & $60.1\%$ \\
  \hline
  \end{tabular}
  }
\end{table}

\subsection{Compatibility to early fusion of features}

Concatenating low-level features has been demonstrated to be powerful in person re-id application \cite{Zhao2014Learning,Zhao2013Unsupervised}. Our approaches can be essentially regarded as a late fusion paradigm where base metrics are first learnt from training samples, which are then fed into visual features to seek their individual weights.
In this section, we study the compatible property of our method to early fusion of visual features. Since we have five types of hand-crafted features, hence there are $\sum_{k=1}^5 C^k_5$ combinatorial strategies in total, as shown in Table \ref{tab:com_feature},  where
\checkmark
denotes the selection of candidate features in concatenation. For instance, $F6$ indicates a feature which is formed by concatenating $F1$ (LAB) and $F2$ (LBP) into a single feature vector. Our approaches are conducted to learn weights for ensembling these base metrics. Evaluation results are reported in Table \ref{tab:early_VIPeR_iLIDS}$-$\ref{tab:early_CUHK01_CUHK03}, where  kLFDA is employed to perform metric learning. We have the following observations:
\begin{itemize}
\item Our ensemble-based approaches consistently outperform any kind of
  concatenated features in rank-$k$ recognition rates, demonstrating the effectiveness in combining these features with adaptive weights. Simply concatenating features is essentially to assign equally pre-defined weights to their metric, which is not robust enough against complex variations in illumination, pose, viewpoints, camera setting, and background clutter across camera views.
\item Incorporating more base metrics is beneficial to our algorithms. In most cases, the increasingly availability of base metrics contribute to either achieve slight improvement in rank-$k$ recognition rate or keep the state-of-the art results from Table \ref{tab:best2}. This demonstrates the robustness of our approaches against feature selection.
\item Some low-level features are complementary to each other, and consequently, their concatenations give promising results, e.g., in iLIDS, F22 (concatenation of LBP, HS, and RGB histogram) shows superior performance to other fusion strategies.  However, this pre-defined combination is lack of generalization, and only suits to specific data sets, e.g., in 3DPeS, F22 is inferior to F25.
\end{itemize}

\begin{table}[h!]\tiny
\centering
\caption{The indication table for early fusion strategies of low-level features. LAB (F1), LBP (F2), HS (F3), RGB (F4), SIFT(F5). }\label{tab:com_feature}
{
\begin{tabular}{c|ccccc||c|ccccc}
\hline\hline
& F1 & F2 & F3 & F4 & F5 & & F1 & F2 & F3 & F4 & F5\\
\hline
F6  & \cellcolor[gray]{.95}\checkmark& \cellcolor[gray]{0.95}\checkmark&  & &  & F19  & \cellcolor[gray]{0.95}\checkmark&  & \cellcolor[gray]{0.95}\checkmark& \cellcolor[gray]{0.95}\checkmark&\\
F7  & \cellcolor[gray]{0.95}\checkmark&  & \cellcolor[gray]{0.95}\checkmark& &  &F20  & \cellcolor[gray]{0.95}\checkmark&  & \cellcolor[gray]{0.95}\checkmark& & \cellcolor[gray]{0.95}\checkmark\\
F8  & \cellcolor[gray]{0.95}\checkmark&  &  & \cellcolor[gray]{0.95}\checkmark& & F21  & \cellcolor[gray]{0.95}\checkmark&  &  & \cellcolor[gray]{0.95}\checkmark& \cellcolor[gray]{0.95}\checkmark\\
F9  & \cellcolor[gray]{0.95}\checkmark&  &  & & \cellcolor[gray]{0.95}\checkmark& F22  &  & \cellcolor[gray]{0.95}\checkmark& \cellcolor[gray]{0.95}\checkmark& \cellcolor[gray]{0.95}\checkmark&\\
F10  &  & \cellcolor[gray]{0.95}\checkmark& \cellcolor[gray]{0.95}\checkmark&  &  & F23  &  & \cellcolor[gray]{0.95}\checkmark& \cellcolor[gray]{0.95}\checkmark&  & \cellcolor[gray]{0.95}\checkmark\\
F11  &  &  \cellcolor[gray]{0.95}\checkmark&  & \cellcolor[gray]{0.95}\checkmark& & F24  &  & \cellcolor[gray]{0.95}\checkmark&  & \cellcolor[gray]{0.95}\checkmark& \cellcolor[gray]{0.95}\checkmark\\
F12  &  & \cellcolor[gray]{0.95}\checkmark&  & & \cellcolor[gray]{0.95}\checkmark&  F25  &  &  & \cellcolor[gray]{0.95}\checkmark& \cellcolor[gray]{0.95}\checkmark& \cellcolor[gray]{0.95}\checkmark\\
F13  &  &  & \cellcolor[gray]{0.95}\checkmark& \cellcolor[gray]{0.95}\checkmark& & F26  & \cellcolor[gray]{0.95}\checkmark& \cellcolor[gray]{0.95}\checkmark& \cellcolor[gray]{0.95}\checkmark& \cellcolor[gray]{0.95}\checkmark&\\
F14  &  &  & \cellcolor[gray]{0.95}\checkmark& & \cellcolor[gray]{0.95}\checkmark& F27  & \cellcolor[gray]{0.95}\checkmark&\cellcolor[gray]{0.95}\checkmark& \cellcolor[gray]{0.95}\checkmark& & \cellcolor[gray]{0.95}\checkmark\\
F15  &  &  &  & \cellcolor[gray]{0.95}\checkmark& \cellcolor[gray]{0.95}\checkmark& F28  & \cellcolor[gray]{0.95}\checkmark& \cellcolor[gray]{0.95}\checkmark&  & \cellcolor[gray]{0.95}\checkmark& \cellcolor[gray]{0.95}\checkmark\\
F16  & \cellcolor[gray]{0.95}\checkmark& \cellcolor[gray]{0.95}\checkmark& \cellcolor[gray]{0.95}\checkmark&  & & F29  & \cellcolor[gray]{0.95}\checkmark&  & \cellcolor[gray]{0.95}\checkmark& \cellcolor[gray]{0.95}\checkmark& \cellcolor[gray]{0.95}\checkmark\\
F17  &  \cellcolor[gray]{0.95}\checkmark& \cellcolor[gray]{0.95}\checkmark&  & \cellcolor[gray]{0.95}\checkmark& & F30  &  & \cellcolor[gray]{0.95}\checkmark& \cellcolor[gray]{0.95}\checkmark& \cellcolor[gray]{0.95}\checkmark& \cellcolor[gray]{0.95}\checkmark\\
F18  & \cellcolor[gray]{0.95}\checkmark&  \cellcolor[gray]{0.95}\checkmark&  & & \cellcolor[gray]{0.95}\checkmark& F31  & \cellcolor[gray]{0.95}\checkmark& \cellcolor[gray]{0.95}\checkmark& \cellcolor[gray]{0.95}\checkmark& \cellcolor[gray]{0.95}\checkmark& \cellcolor[gray]{0.95}\checkmark\\
\hline
\hline
\end{tabular}
}
\end{table}

\begin{table}[t]\scriptsize
  \caption{Rank-1, Rank-5, and Rank-10 recognition rate of various early fusion strategies over VIPeR and iLIDS databases.}  \label{tab:early_VIPeR_iLIDS}
  {
  \begin{tabular}{c|c|c|c||c|c|c}
  \hline
&\multicolumn{3}{|c||}{VIPeR} &  \multicolumn{3}{|c}{iLIDS}\\
\hline
    Feature  & CMC(1)  &  CMC(5) & CMC(10)  & CMC(1)  &  CMC(5) & CMC(10) \\
  \hline\hline
   F1  & $22.4\%$ & $50.6\%$ &67.7\% & $47.4\%$ & $67.7\%$ &81.3\%\\
   F2  &  $35.4\%$ &  $63.6\%$ & 77.2\% &  $59.3\%$ &  $67.7\%$ & 81.3\%\\
   F3  & $23.7\%$ & $54.4\%$ & 70.2\% & $50.8\%$ & $72.8\%$ & 72.8\%\\
   F4  &  $26.2\%$ &  $53.4\%$ & 69.3\% &  $52.5\%$ &  $71.1\%$ & 81.3\%\\
   F5  & $27.8\%$ & $54.7\%$ & 68.6\% & $47.4\%$ & $59.3\%$ & 76.2\%\\
   F6  &  $31.0\%$ &  $63.2\%$ & 78.1\% &  $49.1\%$ &  $71.1\%$ & 83.0\%\\
   F7  & $29.1\%$ & $61.0\%$ & 76.8\% & $50.8\%$ & $69.4\%$ & 79.6\%\\
   F8  &  $25.6\%$ &  $60.7\%$ & 76.5\% &  $54.2\%$ &  $69.4\%$ & 79.6\%\\
   F9  & $34.4\%$ & $68.6\%$ & 80.0\% & $54.2\%$ & $71.1\%$ & 74.5\%\\
   F10  &  $39.5\%$ &  $69.6\%$ & 82.9\% &  $60.7\%$ &  $77.9\%$ & 84.7\%\\
   F11  & $38.6\%$ & $70.2\%$ & 84.8\% & $59.3\%$ & $72.8\%$ & 86.4\%\\
   F12  &  $35.7\%$ &  $68.9\%$ & 82.5\% &  $55.9\%$ &  $66.1\%$ & 77.9\%\\
   F13  & $34.1\%$ & $68.3\%$ & 82.2\% & $52.5\%$ & $71.1\%$ & 74.5\%\\
   F14  &  $38.9\%$ &  $68.6\%$ & 81.3\% &  $50.8\%$ &  $69.4\%$ & 79.6\%\\
   F15  & $37.6\%$ & $71.2\%$ & 83.5\% & $50.8\%$ & $62.7\%$ & 79.6\%\\
\hline
   F16  &  $34.8\%$ &  $68.9\%$ & 81.9\% &  $54.2\%$ &  $72.8\%$ & 84.7\%\\
   F17  & $31.6\%$ & $70.2\%$ & 82.2\% & $54.2\%$ & $71.1\%$ & 83.0\%\\
   F18  &  $38.9\%$ &  $72.1\%$ & 84.4\% &  $55.9\%$ &  $71.1\%$ & 79.6\%\\
   F19  & $31.9\%$ & $66.4\%$ & 81.6\% & $57.6\%$ & $71.2\%$ & 83.0\%\\
   F20  &  $37.3\%$ &  $73.1\%$ &81.6 \% &  $57.6\%$ &  $71.1\%$ & 76.2\%\\
   F21  & $35.7\%$ & $71.2\%$ & 82.9\% & $55.9\%$ & $71.1\%$ & 76.2\%\\
   F22  &  $40.1\%$ &  $75.3\%$ & 86.0\% &  $60.7\%$ &  $74.5\%$ & 83.0\%\\
   F23  & $40.8\%$ & $74.0\%$ & 84.4\% & $59.3\%$ & $67.7\%$ & 83.0\%\\
   F24  &  $43.3\%$ &  $75.3\%$ & 85.4\% &  $54.2\%$ &  $66.1\%$ & 79.6\%\\
   F25  &  $41.4\%$ &  $74.3\%$ & 84.8\% &  $54.2\%$ &  $72.8\%$ & 76.2\%\\
\hline
   F26  & $34.1\%$ & $72.1\%$ & 84.4\% & $57.6\%$ & $71.2\%$ & 84.7\%\\
   F27  &  $40.1\%$ &  $75.6\%$ & 86.0\% &  $57.6\%$ &  $72.8\%$ & 79.6\%\\
   F28  & $39.2\%$ & $73.4\%$ & 86.0\% & $54.2\%$ & $72.8\%$ & 77.9\%\\
   F29  &  $38.9\%$ &  $74.0\%$ & 84.8\% &  $55.9\%$ &  $71.1\%$ & 76.2\%\\
   F30  & $45.8\%$ & $75.9\%$ & 88.2\% & $55.9\%$ & $69.4\%$ & 83.0\%\\
   F31  &  $41.7\%$ &  $75.6\%$ & 87.0\% &  $57.6\%$ &  $72.8\%$ & 79.6\%\\
\hline
   \CMCstruct   &  $\mathbf{47.1\%}$ &  $\mathbf{79.7\%}$ & $\mathbf{88.6\%}$ &  $\mathbf{61.0\%}$ &  $\mathbf{76.3\%}$ & $\mathbf{86.4\%}$\\
   \CMCtriplet  &  $\mathbf{47.7\%}$ & $\mathbf{79.1\%}$ & $\mathbf{88.9\%}$ &  $\mathbf{61.3\%}$ & $\mathbf{77.1\%}$ & $\mathbf{86.7\%}$\\
  \hline
  \end{tabular}
  }
\end{table}

\begin{table}[t]\scriptsize
  \caption{Rank-1, Rank-5, and Rank-10 recognition rate of various early fusion strategies over PRID2011 and 3DPeS databases.
  }  \label{tab:early_PRID2011_3DPeS}
  {
  \begin{tabular}{c|c|c|c||c|c|c}
  \hline
&\multicolumn{3}{|c||}{PRID 2011} &  \multicolumn{3}{|c}{3DPeS}\\
\hline
    Feature  & CMC(1)  &  CMC(5) & CMC(10)  & CMC(1)  &  CMC(5) & CMC(10) \\
  \hline\hline
   F1  & $10.1\%$ & $24.0\%$ &31.0\% &$39.5\%$ & $65.6\%$ & 77.1\%\\
   F2  &  $5.0\%$ &  $17.0\%$ & 29.0\% & $38.5\%$ &  $54.1\%$ &  67.7\%\\
   F3  & $13.0\%$ & $29.0\%$ & 32.0\% & $37.5\%$ & $54.1\%$ &  66.7\%\\
   F4  &  $7.0\%$ &  $18.0\%$ & 24.0\% & $38.5\%$ &  $58.3\%$  &  73.9\%\\
   F5  & $6.0\%$ & $21.0\%$ & 28.0\% & $41.6\%$ & $65.6\%$ & 81.2\%\\
   F6  &  $10.0\%$ &  $28.0\%$ & 36.0\% & 40.6\% & $68.7\%$ &  $76.0\%$\\
   F7  & $11.0\%$ & $28.0\%$ & 35.0\% & 43.7\% &$67.7\%$ & $78.1\%$\\
   F8  &  $11.0\%$ &  $28.0\%$ & 33.0\% &  39.5\% &$65.6\%$ &  $79.1\%$ \\
   F9  & $9.0\%$ & $29.0\%$ & 38.0\%  & 48.9\% &$71.8\%$ & $81.2\%$\\
   F10  &  $11.1\%$ &  $24.0\%$ & 34.0\%  &  40.6\% &$59.3\%$ &  $72.9\%$\\
   F11  & $8.0\%$ & $17.0\%$ & 30.0\% &  38.5\% &$56.2\%$ & $73.9\%$\\
   F12  &  $7.0\%$ &  $22.0\%$ & 35.0\% &  42.7\% &$70.8\%$ &  $84.3\%$\\
   F13  & $15.0\%$ & $33.0\%$ & 36.0\%  & 38.5\% &$64.5\%$ & $72.9\%$\\
   F14  &  $7.0\%$ &  $21.0\%$ & 33.0\%  &  48.9\% &$73.9\%$ &  $80.2\%$\\
   F15  & $10.0\%$ & $23.0\%$ & 32.0\%  &  48.9\% &$76.0\%$ & $84.4\%$\\
\hline
   F16  &  $14.0\%$ &  $28.0\%$ & 35.0\%  &  44.7\% &$68.7\%$ &  $78.1\%$ \\
   F17  & $9.0\%$ & $29.0\%$ & 36.0\%   & 43.7\% &$67.7\%$ & $75.0\%$\\
   F18  &  $11.0\%$ &  $31.0\%$ & 35.0\%  &  44.7\% &$71.8\%$ &  $82.2\%$\\
   F19  & $14.0\%$ & $33.0\%$ & 39.0\%  & 40.6\% &$64.5\%$ & $76.0\%$\\
   F20  &  $9.0\%$ &  $32.0\%$ & 41.0\% &  47.9\% & $70.8\%$ &  $79.1\%$ \\
   F21  & $9.0\%$ & $28.0\%$ & 39.0\%  & 47.9\% & $72.9\%$ & $81.2\%$\\
   F22  &  $13.0\%$ &  $26.0\%$ & 37.0\%  &  39.5\% & $60.4\%$ &  $77.0\%$\\
   F23  & $11.0\%$ & $24.0\%$ & 34.0\%  & 46.8\% & $77.0\%$ & $81.2\%$\\
   F24  &  $11.0\%$ &  $25.0\%$ & 37.0\%  &  45.8\% & $75.0\%$ &  $82.2\%$\\
   F25  &  $8.0\%$ &  $21.0\%$ & 32.0\%  &  51.0\% & $73.9\%$ &  $80.2\%$\\
\hline
   F26  & $13.0\%$ & $30.0\%$ & 37.0\% & 43.7\% & $68.7\%$ & $77.0\%$ \\
   F27  &  $14.0\%$ &  $30.0\%$ & 39.0\% &  46.8\% &$70.8\%$ &  $81.2\%$\\
   F28  & $10.0\%$ & $31.0\%$ & 36.0\%  & 46.8\% & $75.0\%$ & $82.2\%$\\
   F29  &  $14.0\%$ &  $32.0\%$ & 40.0\%  &  48.9\% & $70.8\%$ &  $80.2\%$\\
   F30  & $14.0\%$ & $27.0\%$ & 33.0\%  & 46.8\% & $72.9\%$ & $81.2\%$\\
   F31  &  $14.0\%$ &  $28.0\%$ & 37.0\%  &  46.8\% & $69.7\%$ &  $82.2\%$\\
\hline
   \CMCstruct   &  $\mathbf{21.0\%}$ &  $\mathbf{41.0\%}$ & $\mathbf{49.0\%}$ &  $\mathbf{50.0\%}$ &  $\mathbf{79.1\%}$ & $\mathbf{85.4\%}$\\
   \CMCtriplet  &  $\mathbf{20.0\%}$ & $\mathbf{42.0\%}$ & $\mathbf{50.0\%}$ &  $\mathbf{47.9\%}$ & $\mathbf{72.9\%}$ &$ \mathbf{83.2\%}$\\
  \hline
  \end{tabular}
  }
\end{table}

\begin{table}[t]\scriptsize
  \caption{Rank-1, Rank-5, and Rank-10 recognition rate of various early fusion strategies over CUHK01 and CUHK03 databases. }\label{tab:early_CUHK01_CUHK03}
  {
  \begin{tabular}{c|c|c|c||c|c|c}
  \hline
&\multicolumn{3}{|c||}{CUHK01} &  \multicolumn{3}{|c}{CUHK03}\\
\hline
    Feature  & CMC(1)  &  CMC(5) & CMC(10)  & CMC(1)  &  CMC(5) & CMC(10) \\
  \hline\hline
   F1  & $18.7\%$ & $40.0\%$ & 51.9\% & $30.0\%$ & $60.0\%$ & 72.0\%\\
   F2  &  $33.8\%$ &  $59.3\%$ & 70.5\% &  $60.0\%$ &  $84.0\%$ & 92.0\%\\
   F3  & $21.0\%$ & $37.9\%$ & 47.4\% & $27.0\%$ & $61.0\%$ & 68.0\%\\
   F4  &  $14.6\%$ &  $30.9\%$ & 40.8\% &  $23.0\%$ &  $43.0\%$ & 63.0\%\\
   F5  & $36.9\%$ & $58.9\%$ & 68.4\% & $27.0\%$ & $55.0\%$ & 65.0\%\\
   F6  &  $25.5\%$ &  $54.6\%$ & 67.2\% &  $53.0\%$ &  $78.0\%$ & 86.0\%\\
   F7  & $24.7\%$ & $49.2\%$ & 61.8\% & $47.0\%$ & $73.0\%$ & 82.0\%\\
   F8  &  $25.3\%$ &  $50.7\%$ & 60.2\% &  $39.0\%$ &  $68.0\%$ & 76.0\%\\
   F9  & $36.9\%$ & $63.1\%$ & 75.4\% & $36.0\%$ & $66.0\%$ & 78.0\%\\
   F10  &  $39.2\%$ &  $61.4\%$ & 71.3\% &  $52.0\%$ &  $83.0\%$ & 89.0\%\\
   F11  & $34.0\%$ & $57.1\%$ & 69.4\% & $51.0\%$ & $85.0\%$ & 92.0\%\\
   F12  &  $45.3\%$ &  $72.1\%$ & 81.4\% &  $47.0\%$ &  $77.0\%$ & 85.0\%\\
   F13  & $27.8\%$ & $49.4\%$ & 58.2\% & $37.0\%$ & $71.0\%$ & 82.0\%\\
   F14  &  $41.2\%$ &  $63.9\%$ & 74.0\% &  $38.0\%$ &  $68.0\%$ & 81.0\%\\
   F15  & $42.2\%$ & $68.6\%$ & 76.1\% & $37.0\%$ & $64.0\%$ & 82.0\%\\
\hline
   F16  &  $32.1\%$ &  $58.7\%$ & 71.9\% &  $55.0\%$ &  $84.0\%$ & 88.0\%\\
   F17  & $31.5\%$ & $60.2\%$ & 70.7\% & $52.0\%$ & $83.0\%$ & 87.0\%\\
   F18  &  $42.8\%$ &  $69.9\%$ & 81.4\% &  $46.0\%$ &  $78.0\%$ & 86.0\%\\
   F19  & $28.6\%$ & $54.0\%$ & 66.2\% & $45.0\%$ & $75.0\%$ & 85.0\%\\
   F20  &  $41.0\%$ &  $66.8\%$ & 77.5\% &  $47.0\%$ &  $76.0\%$ & 85.0\%\\
   F21  & $39.1\%$ & $67.8\%$ & 77.1\% & $40.0\%$ & $72.0\%$ & 84.0\%\\
   F22  &  $37.1\%$ &  $63.3\%$ & 72.8\% &  $59.0\%$ &  $84.0\%$ & 92.0\%\\
   F23  & $49.9\%$ & $74.0\%$ & 81.8\% & $52.0\%$ & $83.0\%$ & 90.0\%\\
   F24  &  $47.8\%$ &  $75.2\%$ & 81.0\% &  $50.0\%$ &  $80.0\%$ & 91.0\%\\
   F25  &  $45.3\%$ &  $70.3\%$ & 77.7\% &  $42.0\%$ &  $74.0\%$ & 86.0\%\\
\hline
   F26  & $34.0\%$ & $63.7\%$ & 72.5\% & $60.0\%$ & $83.0\%$ & 90.0\%\\
   F27  &  $46.4\%$ &  $72.1\%$ & 80.8\% &  $55.0\%$ &  $82.0\%$ & 91.0\%\\
   F28  & $44.7\%$ & $71.9\%$ & 81.4\% & $52.0\%$ & $82.0\%$ & 89.0\%\\
   F29  &  $42.0\%$ &  $68.8\%$ & 78.1\% &  $51.0\%$ &  $78.0\%$ & 88.0\%\\
   F30  & $49.2\%$ & $75.9\%$ & 81.8\% & $54.0\%$ & $85.0\%$ & 92.0\%\\
   F31  &  $46.5\%$ &  $72.7\%$ & 80.6\% &  $55.0\%$ &  $84.0\%$ & 90.0\%\\
\hline
 {\CMCstruct}   &  $\mathbf{57.1\%}$ &  $\mathbf{76.9\%}$ & $\mathbf{84.9 \%}$  &  $\mathbf{68.0\%}$ &  $\mathbf{86.0\%}$ & $\mathbf{93.0\%}$\\
{\CMCtriplet}  &  $\mathbf{57.5\%}$ & $\mathbf{77.7\%}$ & $\mathbf{84.3\%}$ &  $\mathbf{69.0\%}$ & $\mathbf{87.0\%}$ & $\mathbf{92.0\%}$\\
  \hline
  \end{tabular}
  }
\end{table}

\subsection{Comparison with metric learning algorithms}

We evaluate the performance on different metric learning algorithms: large scale metric learning from equivalence constraints (KISSME) \cite{Kostinger2012Large} \footnote{\url{http://lrs.icg.tugraz.at/research/kissme/}}, large margin nearest neighbor (LMNN) \cite{Weinberger2006Distance} \footnote{\url{http://www.cse.wustl.edu/~kilian/code/lmnn/lmnn.html}} and logistic distance metric learning (LDML) \cite{Guillaumin2009Isthatyou} \footnote{\url{http://lear.inrialpes.fr/people/guillaumin/code.php}}.

\begin{table*}[h]\footnotesize
  \centering
  \caption{Comparison with different metric learning algorithms against varied PCA dimensions. } \label{tab:compare_metric}
  {
  \begin{tabular}{l|c|c|c}
  \hline
    Algorithm  & VIPeR & iLIDS & CUHK01  \\
  \hline
  \hline
   KISSME (32,100) & $34.8\% (28.5\%,28.6\%)$ & $53.3\% (50.0\%,53.3\%)$ & $25.5\% (19.3\%,25.5\%)$ \\
   LMNN (32,100)  & $28.2\% (24.7\%,28.1\%)$ & $42.9\% (39.8\%42.9\%)$ & $24.0\% (18.7\%,24.2\%)$ \\
   LDML (32,100)  & $27.8\% (24.1\%,28.0\%)$ & $42.4\% (38.0\%,42.3\%)$ & $23.9\% (18.1\%,23.8\%)$ \\
  \CMCstruct (32,100) & 35.4\% (33.2\%,35.4\%) & 67.1\% (54.1\%,67.2\%) & 30.7\% (26.4\%,30.7\%) \\
  \hline
  \end{tabular}
  }
\end{table*}

\paragraph{Metric learning}
We first apply principal component analysis (PCA) to reduce the dimensionality and remove noise. Without performing PCA, it is computationally infeasible to perform  metric learning on KISSME. To ensure a fair comparison, we use the same number of PCA dimensions in all metric learning algorithms.
In this experiment, we reduce the feature dimension to 32, 64, and 100 dimensional subspaces. The setting of parameters for these metric learning algorithms are as follows:
For KISSME and LDML, all parameters are set to be the default values as suggested by the authors \cite{Kostinger2012Large,Guillaumin2009Isthatyou}.
In the LMNN algorithm, we keep most parameters to be default values as suggested by the authors \cite{Weinberger2006Distance} while changing the number of nearest neighbors to be 1 and the maximum number of iterations to $1000$ to speed up the experiment. (The default is $10000$).

\paragraph{Experimental results}
We compare rank-$1$ identification rates of three different metric learning algorithms in Table~\ref{tab:compare_metric} (The results with 32-dim and 100-dim are shown in parentheses). The three metric learning algorithms first apply PCA to reduce the dimension of each feature and then take the concatenation as input. \CMCstruct is conducted to learn a set of weights adaptive to these distance functions. We observe that we achieve superior performance to state-of-the-art metric learning methods that combine the same range of feature vectors as input. However the performance of all metric learning algorithms improves as we increase the number of PCA dimensions.  Possible reasons that might have caused significant performance
differences between our obtained results and baselines of metric learning are:
(a) learning weights for distance functions is more adaptive to sample variations in a variety of person re-id databases;  (b) optimizing the test criteria directly is beneficial to improving recognition rate.

\subsection{Approximating kernel learning}\label{ssec:kernel_app}

\begin{figure*}[h!]
    \centering
    \includegraphics[width=0.3\textwidth,clip]{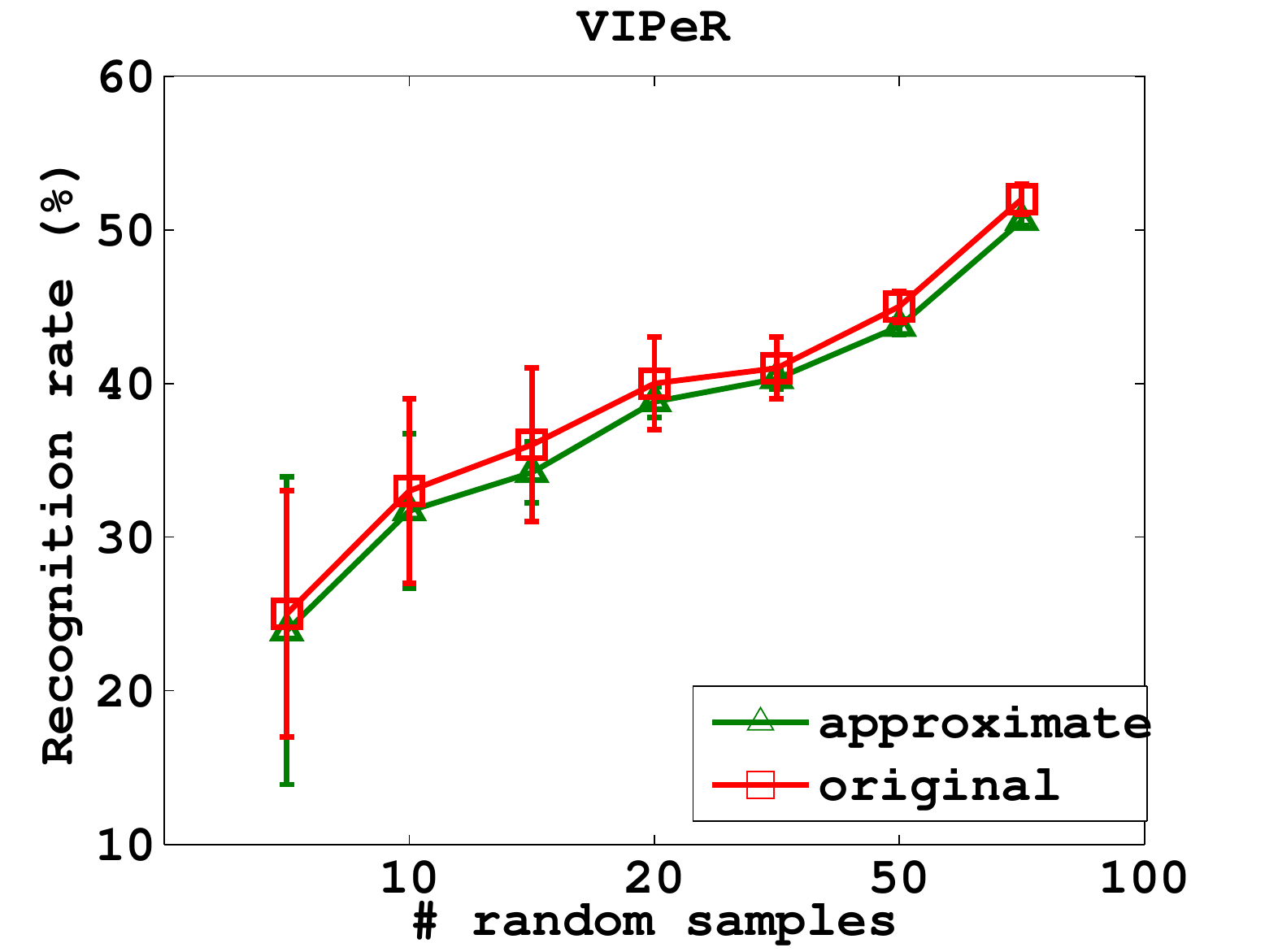}
    \includegraphics[width=0.3\textwidth,clip]{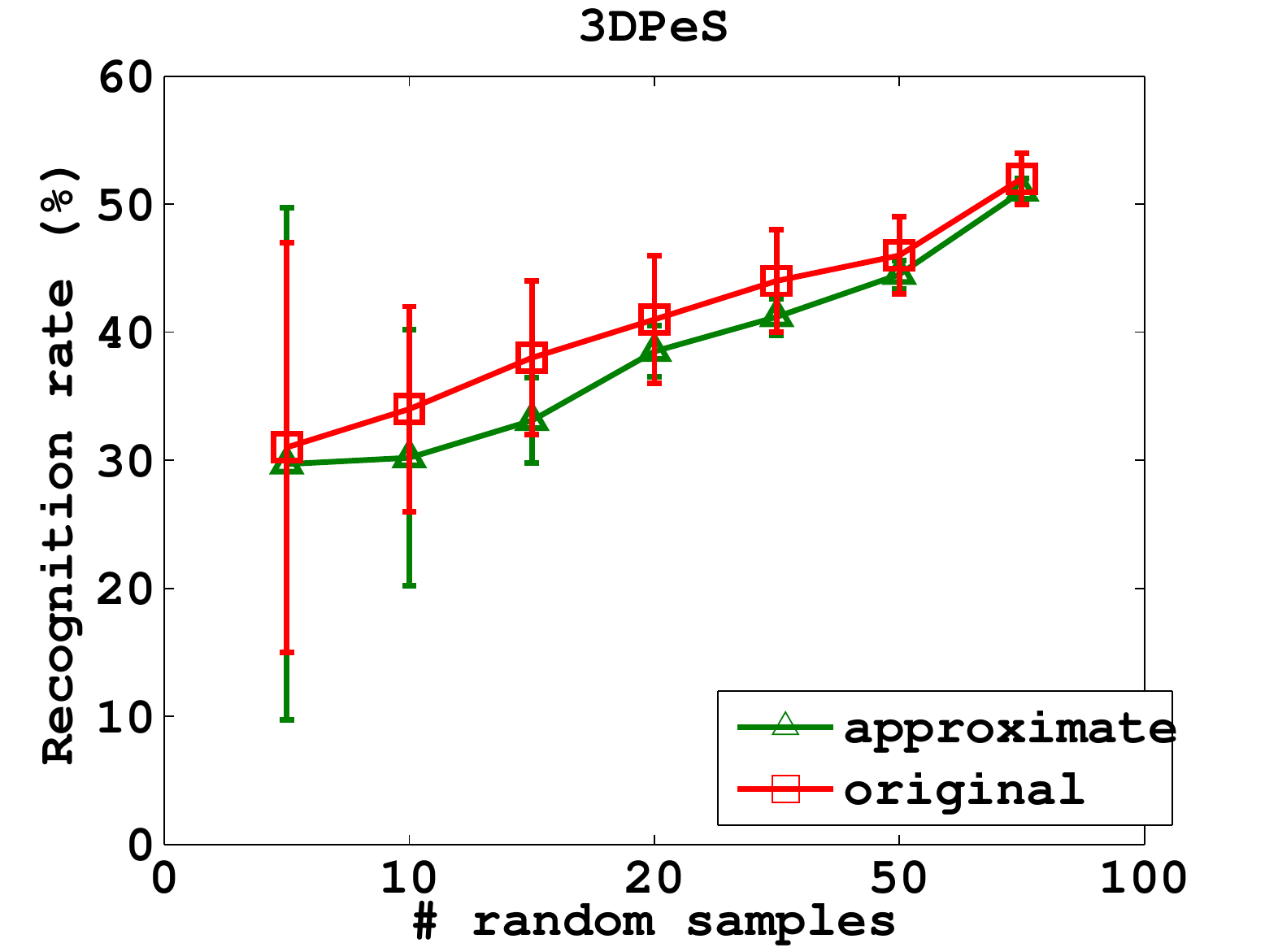}
    \includegraphics[width=0.3\textwidth,clip]{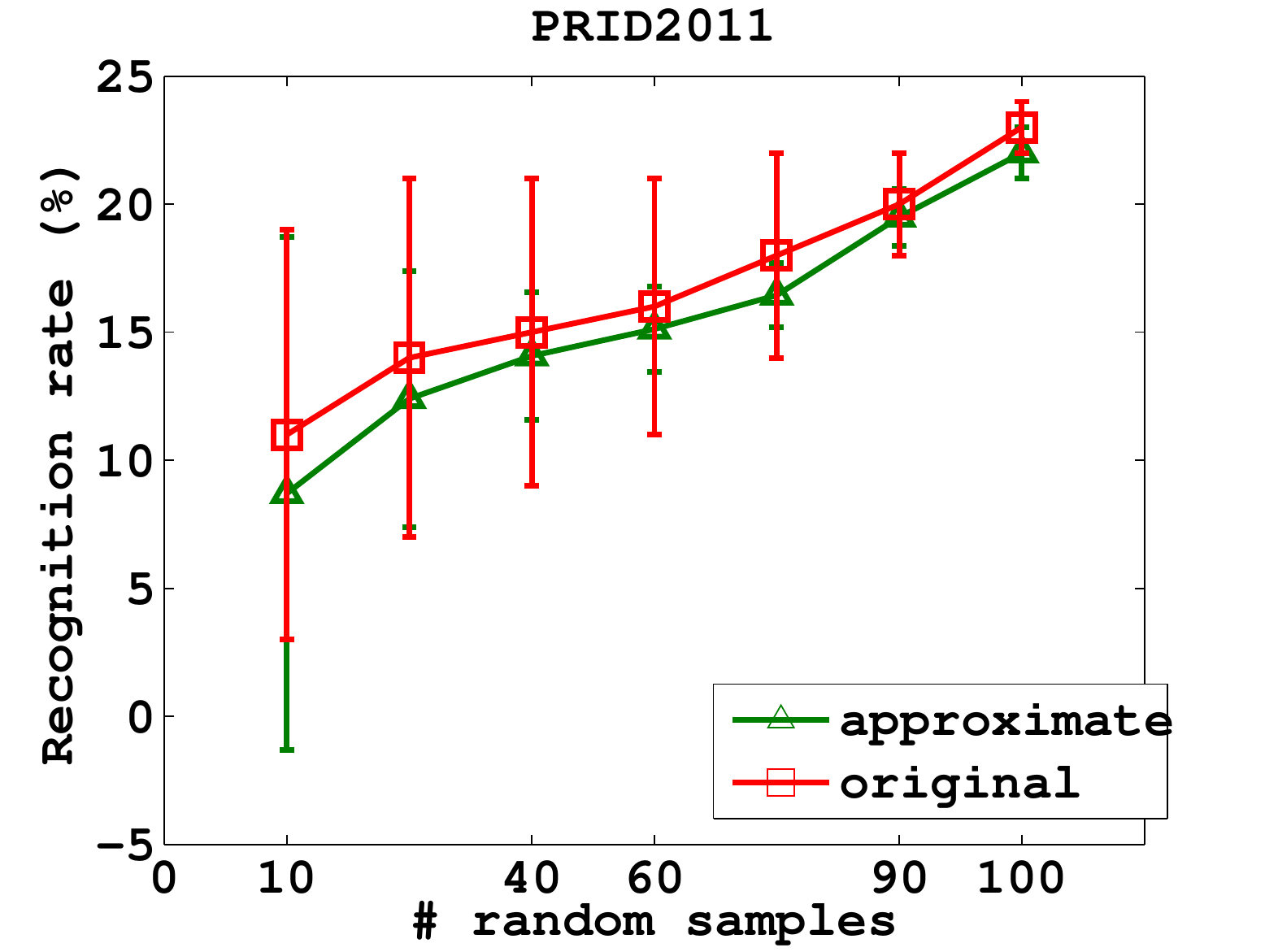}
    \includegraphics[width=0.3\textwidth,clip]{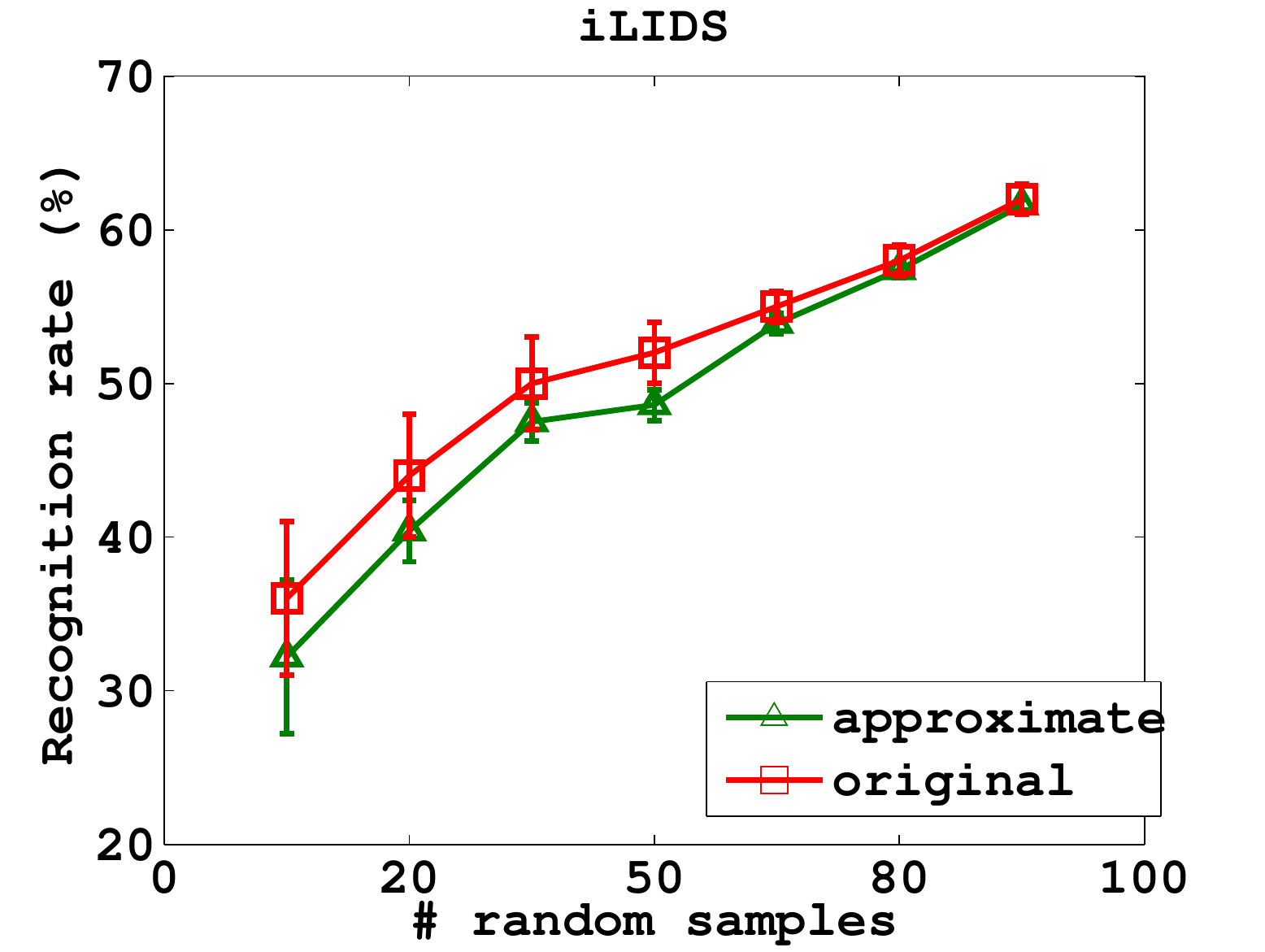}
    \includegraphics[width=0.3\textwidth,clip]{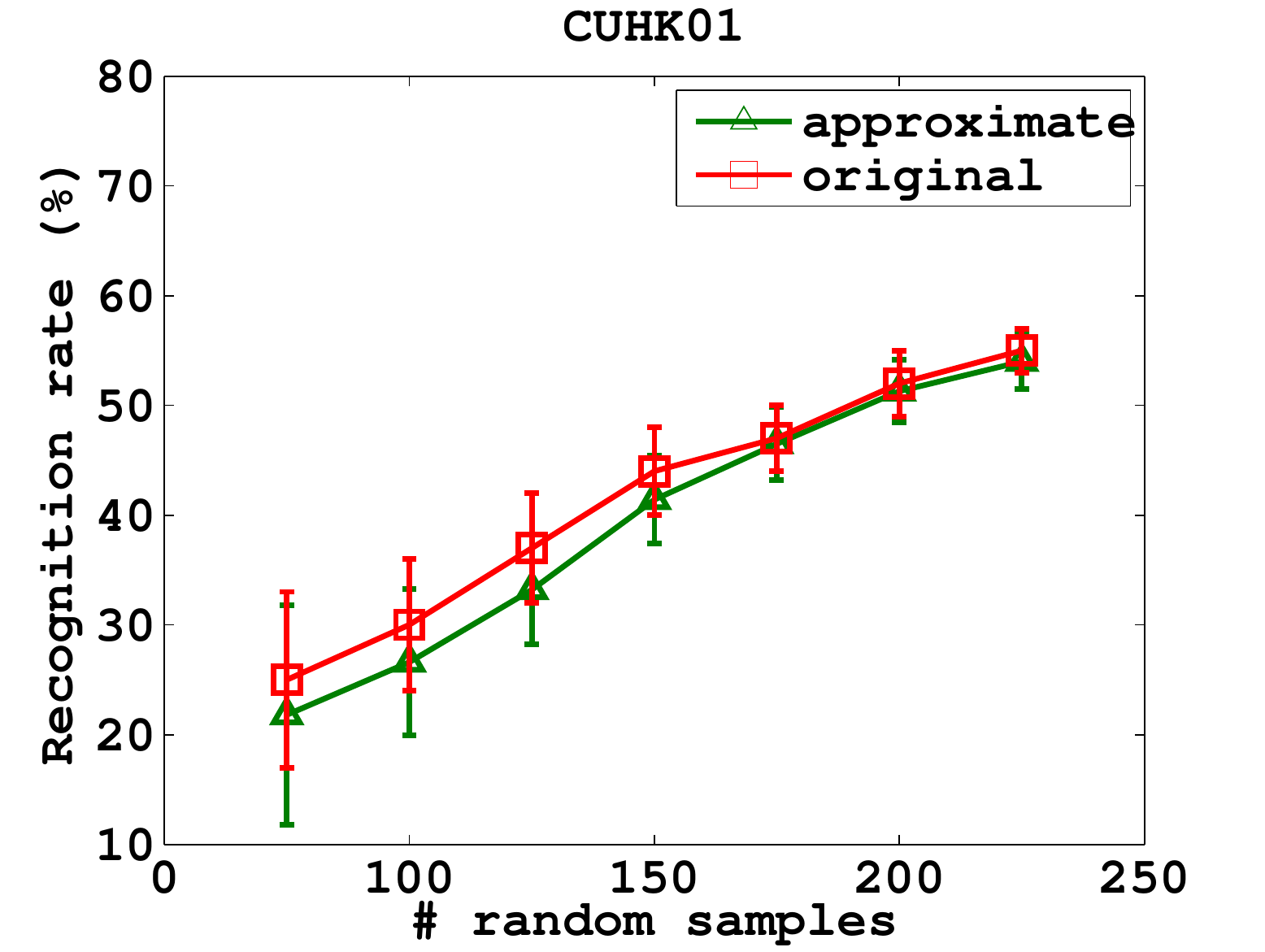}
    \includegraphics[width=0.3\textwidth,clip]{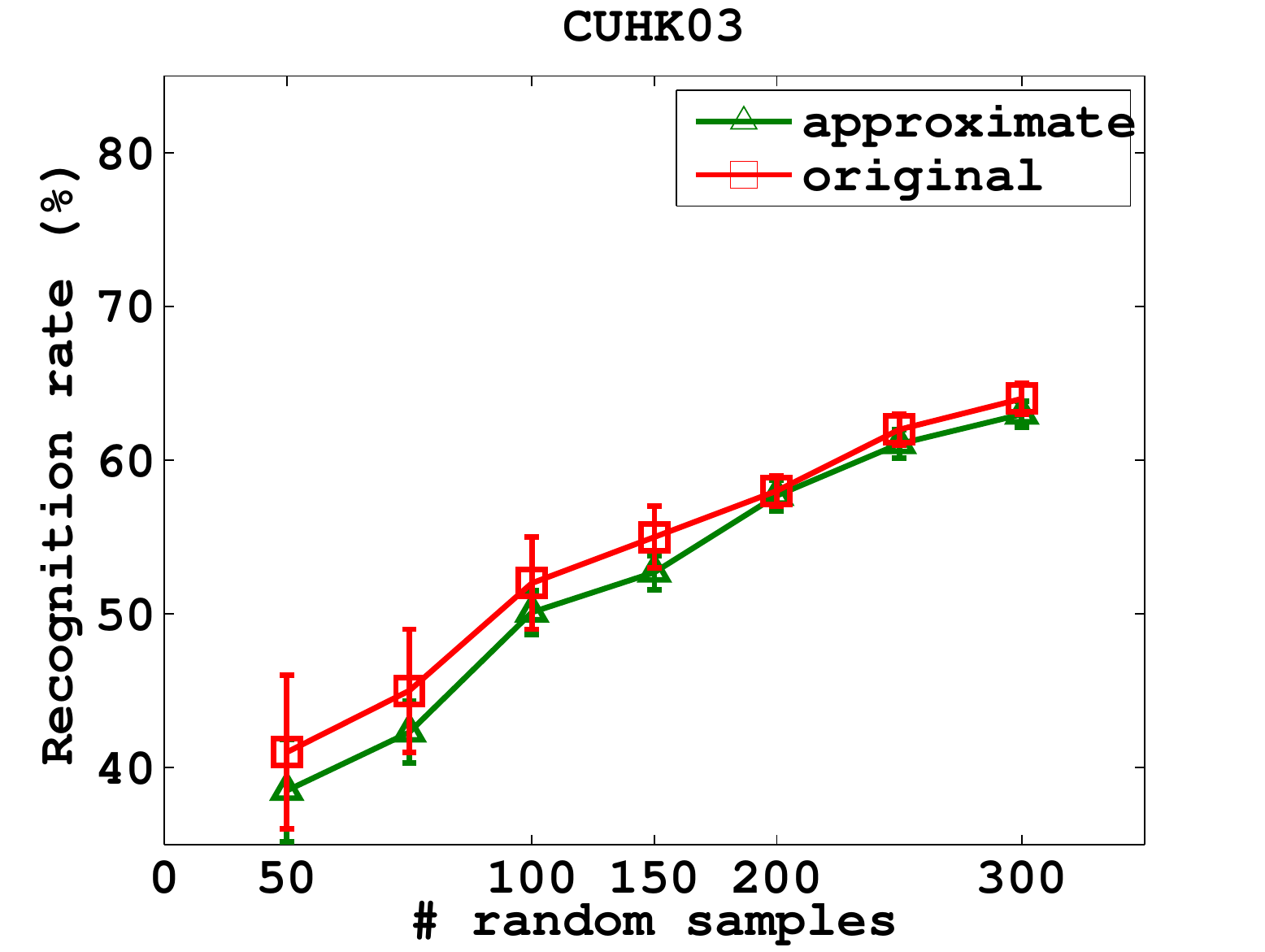}
    \caption{Approximate kernel learning by using the Ny\-s\-tr\-\"{o}\-m method. The recognition rate at rank-1 of \CMCstruct is reported (with std.) against varied number of random samples.The more samples we use, the more accurate results will be. To make the computation efficient, we select less than 300 samples to compute the kernel, which can approximate kernel learning without losing much recognition values compared with the original.} \label{fig:kernel_app}
\end{figure*}

\begin{figure*}[h!]
    \centering
        \includegraphics[width=0.3\textwidth,clip]{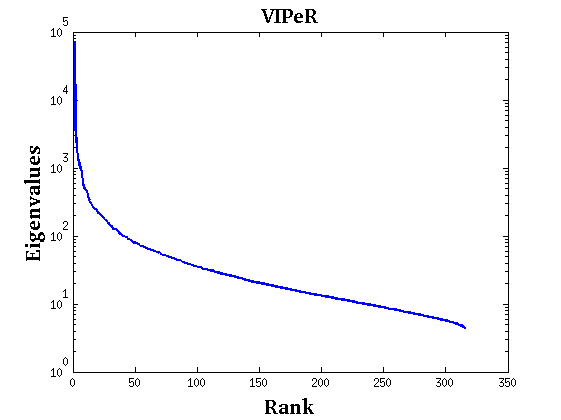}
        \includegraphics[width=0.3\textwidth,clip]{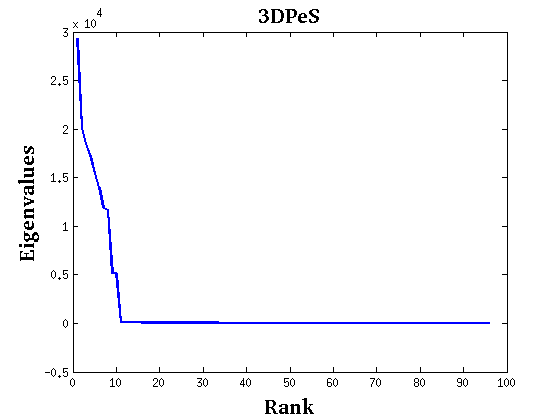}
        \includegraphics[width=0.3\textwidth,clip]{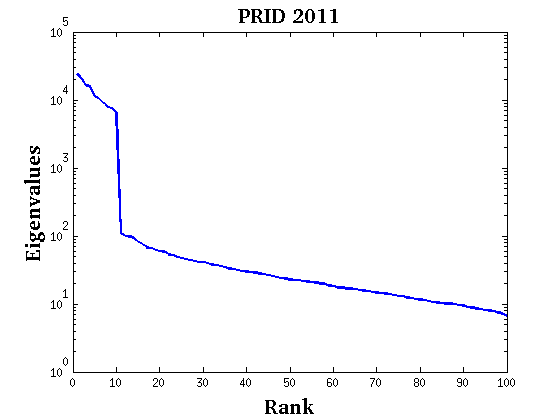}
        \includegraphics[width=0.3\textwidth,clip]{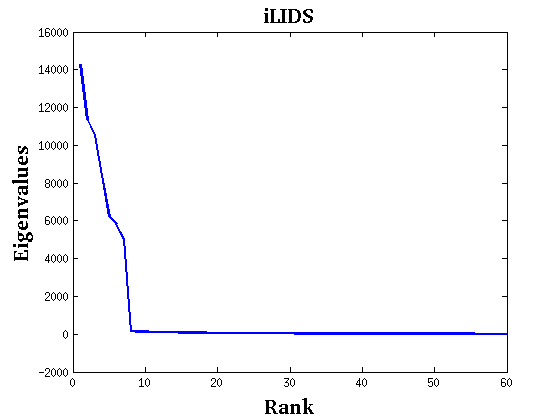}
        \includegraphics[width=0.3\textwidth,clip]{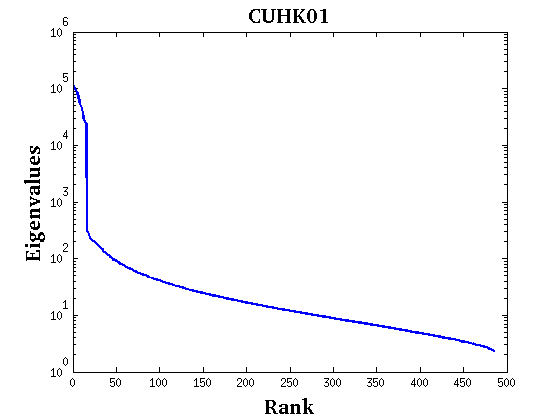}
        \includegraphics[width=0.3\textwidth,clip]{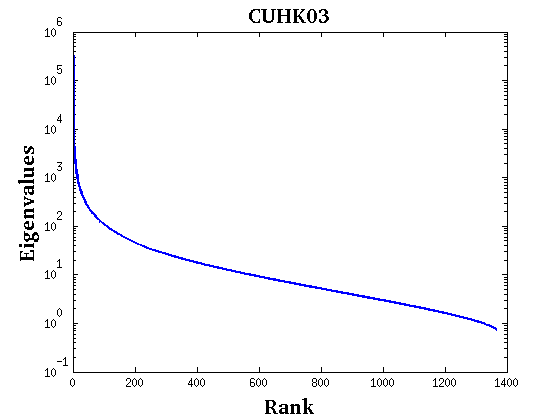}
    \caption{The eigenvalue distributions of kernel matrices. For each dataset, the eigenvalues drop very quickly as the rank increases, leading to a large gap between the top eigenvalue and the remainings. This justifies the feasibility of Ny\-s\-tr\-\"{o}\-m method on person re-id.} \label{fig:eigenvalue}
\end{figure*}

As shown in above empirical study, an ensemble of kernel based metrics remarkably outperforms that of liner base metrics, especially in VIPeR, PRID 2011, CUHK01 and CUHK03. Nonetheless,  the application of non-linear metric learning is recognized to be limited in the case of large-scale data set such as CUHK03 or even medium-sized data set such as CUHK01 due to the high computational cost in kernel learning. To this end, we resort to approximate kernel learning by using the Ny\-s\-tr\-\"{o}\-m method which is able to approximate the full kernel matrix by a low rank matrix with the additional error of $O(1/m)$ in the generalization performance. Fig. \ref{fig:kernel_app} shows the performance of our method with varied number of random samples. Note that for the large data set CUHK03, we restrict the maximum number of random samples to 300 because of the high computational cost. We observe that the approximate counterpart is able to achieve comparable results to the original method with a full kernel matrix. We finally evaluate
whether the key assumption of large eigengap holds for these person re-id data sets. As shown in Fig. \ref{fig:eigenvalue} (eigenvalues are in logarithm scale), it can be seen that the eigenvalues drop very quickly as the rank increases, yielding a dramatic gap between the top eigenvalues and the remaining ones.

\section{Conclusion}\label{sec:con}
In this paper, we have presented two effective structured learning based approaches for person
re-id by combining multiple low-level visual features into a single framework. The approaches are generalized and adaptive to different person re-id datasets by automatically discovering the effects of visual features in recognizing persons. Extensive experimental studies demonstrate the our approaches advance the state-of-the-art results by a significant margin. Our framework is practical to real-world applications since the performance can be concentrated in the range of most practical importance. Moreover our approaches are flexible and can be applied
to any metric learning algorithms. Future works include incorporating depth from a single monocular image \cite{Liu2015Deep}, integrating person re-id with person detector \cite{paisitkriangkrai2008fast} and improving multiple target tracking of \cite{Milan2014Continuous} with  the proposed approaches.

\bibliographystyle{elsarticle-harv}
\bibliography{cviubib}

\begin{thebibliography}{51}
\expandafter\ifx\csname natexlab\endcsname\relax\def\natexlab#1{#1}\fi
\expandafter\ifx\csname url\endcsname\relax
  \def\url#1{\texttt{#1}}\fi
\expandafter\ifx\csname urlprefix\endcsname\relax\def\urlprefix{URL }\fi

\bibitem[{Baltieri et~al.(2011)Baltieri, Vezzani, and
  Cucchiara}]{Baltieri20113dpes}
Baltieri, D., Vezzani, R., Cucchiara, R., 2011. {3DPes}: {3D} people dataset
  for surveillance and forensics. In: Proc. of Int'l. Workshop on Mult. Acc. to
  3D Human Objs.

\bibitem[{Bazzani et~al.(2012)Bazzani, Cristani, Perina, and
  Murino}]{Bazzani2012Multiple}
Bazzani, L., Cristani, M., Perina, A., Murino, V., 2012. Multiple-shot person
  re-identification by chromatic and epitomic analyses. Patt. Recogn. 33~(7),
  898--903.

\bibitem[{Cheng et~al.(2011)Cheng, Cristani, Stoppa, Bazzani, and
  Murino}]{Cheng2011Custom}
Cheng, D.~S., Cristani, M., Stoppa, M., Bazzani, L., Murino, V., 2011. Custom
  pictorial structures for re-identification. In: Proc. British Mach. Vis.
  Conf.

\bibitem[{Chopra et~al.(2005)Chopra, Hadsell, and LeCun}]{Chopra2005Learning}
Chopra, S., Hadsell, R., LeCun, Y., 2005. Learning a similarity metric
  discriminatively, with application to face verification. In: Proc. IEEE Conf.
  Comp. Vis. Patt. Recogn.

\bibitem[{Davis et~al.(2007)Davis, Kulis, Jain, Sra, and
  Dhillon}]{Davis2007Information}
Davis, J.~V., Kulis, B., Jain, P., Sra, S., Dhillon, I.~S., 2007.
  Information-theoretic metric learning. In: Proc. Int. Conf. Mach. Learn.

\bibitem[{Drineas and Mahoney(2005)}]{Nystrom2005JMLR}
Drineas, P., Mahoney, M.~W., 2005. On the system for approximating a gram
  matrix for improved kernel-based leraning. J. Mach. Learn. Res. 6,
  2153--2175.

\bibitem[{Farenzena et~al.(2010)Farenzena, Bazzani, Perina, Murino, and
  Cristani}]{Farenzena2010Person}
Farenzena, M., Bazzani, L., Perina, A., Murino, V., Cristani, M., 2010. Person
  re-identification by symmetry-driven accumulation of local features. In:
  Proc. IEEE Conf. Comp. Vis. Patt. Recogn.

\bibitem[{Felzenszwalb et~al.(2010)Felzenszwalb, Girshick, McAllester, and
  Ramanan}]{Felzenszwalb2010Object}
Felzenszwalb, P., Girshick, R., McAllester, D., Ramanan, D., 2010. Object
  detection with discriminatively trained part based models. {IEEE} Trans.
  Pattern Anal. Mach. Intell. 32~(9), 1627--1645.

\bibitem[{Gheissari et~al.(2006)Gheissari, Sebastian, and
  Hartley}]{Gheissari2006Person}
Gheissari, N., Sebastian, T.~B., Hartley, R., 2006. Person reidentification
  using spatiotemporal appearance. In: Proc. IEEE Conf. Comp. Vis. Patt.
  Recogn.

\bibitem[{Gray et~al.(2007)Gray, Brennan, and Tao}]{Gray2007Evaluating}
Gray, D., Brennan, S., Tao, H., 2007. Evaluating appearance models for
  recognition, reacquisition, and tracking. In: Proc. Int'l. Workshop on Perf.
  Eval. of Track. and Surv'l.

\bibitem[{Gray and Tao(2008)}]{Gray2008Viewpoint}
Gray, D., Tao, H., 2008. Viewpoint invariant pedestrian recognition with an
  ensemble of localized features. In: Proc. Eur. Conf. Comp. Vis.

\bibitem[{Guillaumin et~al.(2009)Guillaumin, Verbeek, and
  Schmid}]{Guillaumin2009Isthatyou}
Guillaumin, M., Verbeek, J., Schmid, C., 2009. Is that you? metric learning
  approaches for face identification. In: Proc. IEEE Int. Conf. Comp. Vis.

\bibitem[{Hirzer et~al.(2011)Hirzer, Beleznai, Roth, and
  Bischof}]{Hirzer2011Person}
Hirzer, M., Beleznai, C., Roth, P.~M., Bischof, H., 2011. Person
  re-identification by descriptive and discriminative classification. In: Proc.
  Scandinavian Conf. on Image Anal.

\bibitem[{Hirzer et~al.(2012)Hirzer, Roth, and Bischof}]{Hirzer2012Person}
Hirzer, M., Roth, P., Bischof, H., 2012. Person re-identification by efficient
  imposter-based metric learning. In: Proc. Int'l. Conf. on Adv. Vid. and Sig.
  Surveillance.

\bibitem[{Joachims(2005)}]{Joachims2005Support}
Joachims, T., 2005. A support vector method for multivariate performance
  measures. In: Proc. Int. Conf. Mach. Learn.

\bibitem[{Kedem et~al.(2012)Kedem, Tyree, Sha, Lanckriet, and
  Weinberger}]{Kedem2012Nonlinear}
Kedem, D., Tyree, S., Sha, F., Lanckriet, G.~R., Weinberger, K.~Q., 2012.
  Non-linear metric learning. In: Proc. Adv. Neural Inf. Process. Syst.

\bibitem[{Kostinger et~al.(2012)Kostinger, Hirzer, Wohlhart, Roth, and
  Bischof}]{Kostinger2012Large}
Kostinger, M., Hirzer, M., Wohlhart, P., Roth, P.~M., Bischof, H., 2012. Large
  scale metric learning from equivalence constraints. In: Proc. IEEE Conf.
  Comp. Vis. Patt. Recogn.

\bibitem[{Li et~al.(2012)Li, Zhao, and Wang}]{Li2012Human}
Li, W., Zhao, R., Wang, X., 2012. Human reidentification with transferred
  metric learning. In: Proc. Asian Conf. Comp. Vis.

\bibitem[{Li et~al.(2014)Li, Zhao, Xiao, and Wang}]{Li2014Deep}
Li, W., Zhao, R., Xiao, T., Wang, X., 2014. Deepreid: Deep filter pairing
  neural network for person re-identification. In: Proc. IEEE Conf. Comp. Vis.
  Patt. Recogn.

\bibitem[{Li et~al.(2013)Li, Chang, Liang, Huang, Cao, and
  Smith}]{Li2013Learning}
Li, Z., Chang, S., Liang, F., Huang, T.~S., Cao, L., Smith, J., 2013. Learning
  locally-adaptive decision functions for person verification. In: Proc. IEEE
  Conf. Comp. Vis. Patt. Recogn.

\bibitem[{Liao et~al.(2015)Liao, Hu, Zhu, and Li}]{XQDA}
Liao, S., Hu, Y., Zhu, X., Li, S.~Z., 2015. Person re-identification by local
  maximal occurrence representation and metric learning. In: Proc. IEEE Conf.
  Comp. Vis. Patt. Recogn.

\bibitem[{Lisanti et~al.(2015)Lisanti, Masi, and Andrew
  D.~Bagdanov}]{Giuseppe2015PAMI}
Lisanti, G., Masi, I., Andrew D.~Bagdanov, A. D.~B., 2015. Person
  re-identification by iterative re-weighted sparse ranking. {IEEE} Trans.
  Pattern Anal. Mach. Intell. PP~(99), 1.

\bibitem[{Liu et~al.(2012)Liu, Gong, Loy, and Lin}]{WhatFeatures}
Liu, C., Gong, S., Loy, C.~C., Lin, X., 2012. Person re-identification: What
  features are important? In: International Workshop on Re-identification,
  ECCV.

\bibitem[{Liu et~al.(2015)Liu, Shen, and Lin}]{Liu2015Deep}
Liu, F., Shen, C., Lin, G., 2015. Deep convolutional neural fields for depth
  estimation from a single image. In: Proc. IEEE Conf. Comp. Vis. Patt. Recogn.

\bibitem[{Lowe(2004)}]{Lowe2004Distinctive}
Lowe, D.~G., 2004. Distinctive image features from scale-invariant keypoints.
  Int. J. Comp. Vis. 60~(2), 91--110.

\bibitem[{McFee and Lanckriet(2010)}]{McFee2010Metric}
McFee, B., Lanckriet, G. R.~G., 2010. Metric learning to rank. In: Proc. Int.
  Conf. Mach. Learn.

\bibitem[{Mignon and Jurie(2012)}]{PCCA}
Mignon, A., Jurie, F., 2012. Pcca: a new approach for distance learning from
  sparse pairwise constraints. In: Proc. IEEE Conf. Comp. Vis. Patt. Recogn.
  pp. 2666--2672.

\bibitem[{Milan et~al.(2014)Milan, Roth, and Schindler}]{Milan2014Continuous}
Milan, A., Roth, S., Schindler, K., 2014. Continuous energy minimization for
  multitarget tracking. {IEEE} Trans. Pattern Anal. Mach. Intell. 36~(1),
  58--72.

\bibitem[{Narasimhan and Agarwal(2013)}]{Narasimhan2013Structural}
Narasimhan, H., Agarwal, S., 2013. A structural svm based approach for
  optimizing partial auc. In: Proc. Int. Conf. Mach. Learn.

\bibitem[{Ojala et~al.(2002)Ojala, Pietikainen, and
  Maenpaa}]{Ojala2002Multiresolution}
Ojala, T., Pietikainen, M., Maenpaa, T., 2002. Multiresolution gray-scale and
  rotation invariant texture classification with local binary patterns. {IEEE}
  Trans. Pattern Anal. Mach. Intell. 24~(7), 971--987.

\bibitem[{Paisitkriangkrai et~al.(2015)Paisitkriangkrai, Shen, and van~den
  Hengel}]{paul2015ensemble}
Paisitkriangkrai, S., Shen, C., van~den Hengel, A., 2015. Learning to rank in
  person re-identification with metric ensembles. In: Proc. IEEE Conf. Comp.
  Vis. Patt. Recogn. Boston, USA.

\bibitem[{Paisitkriangkrai et~al.(2008)Paisitkriangkrai, Shen, and
  Zhang}]{paisitkriangkrai2008fast}
Paisitkriangkrai, S., Shen, C., Zhang, J., 2008. Fast pedestrian detection
  using a cascade of boosted covariance features. {IEEE} Trans. Circuits Syst.
  Video Technol. 18~(8), 1140--1151.

\bibitem[{Pedagadi et~al.(2013)Pedagadi, Orwell, Velastin, and
  Boghossian}]{Pedagadi2013Local}
Pedagadi, S., Orwell, J., Velastin, S., Boghossian, B., 2013. Local fisher
  discriminant analysis for pedestrian re-identification. In: Proc. IEEE Conf.
  Comp. Vis. Patt. Recogn.

\bibitem[{Prosser et~al.(2010)Prosser, Zheng, Gong, Xiang, and
  Mary}]{Prosser2010Person}
Prosser, B., Zheng, W.~S., Gong, S., Xiang, T., Mary, Q., 2010. Person
  re-identification by support vector ranking. In: Proc. British Mach. Vis.
  Conf.

\bibitem[{Rahimi and Recht(2007)}]{nips2007random}
Rahimi, A., Recht, B., 2007. Random features for large-scale kernel machines.
  In: Proc. Adv. Neural Inf. Process. Syst. pp. 1177--1184.

\bibitem[{Roth et~al.(2014)Roth, Hirzer, K{\"o}stinger, Beleznai, and
  Bischof}]{Roth2014Mahalanobis}
Roth, P.~M., Hirzer, M., K{\"o}stinger, M., Beleznai, C., Bischof, H., 2014.
  Mahalanobis distance learning for person re-identification. In: Person
  Re-Identification. Springer, pp. 247--267.

\bibitem[{Schultz and Joachims(2004)}]{Schultz2004Learning}
Schultz, M., Joachims, T., 2004. Learning a distance metric from relative
  comparisons. In: Proc. Adv. Neural Inf. Process. Syst.

\bibitem[{Schwartz and Davis(2009)}]{Schwartz2009Learning}
Schwartz, W., Davis, L., 2009. Learning discriminative appearance-based models
  using partial least squares. In: Proc. of SIBGRAPI.

\bibitem[{Wang et~al.(2007)Wang, Doretto, Sebastian, Rittscher, and
  Tu}]{Wang2007Shape}
Wang, X., Doretto, G., Sebastian, T., Rittscher, J., Tu, P., 2007. Shape and
  appearance context modeling. In: Proc. IEEE Int. Conf. Comp. Vis.

\bibitem[{Wang et~al.(2009)Wang, Han, and Yan}]{Wang2009HOG}
Wang, X., Han, T.~X., Yan, S., 2009. An {HOG-LBP} human detector with partial
  occlusion handling. In: Proc. IEEE Int. Conf. Comp. Vis.

\bibitem[{Weinberger et~al.(2006)Weinberger, Blitzer, and
  Saul}]{Weinberger2006Distance}
Weinberger, K., Blitzer, J., Saul, L., 2006. Distance metric learning for large
  margin nearest neighbor classification. In: NIPS.

\bibitem[{Weinberger and Saul(2008)}]{Weinberger2008Fast}
Weinberger, K.~Q., Saul, L.~K., 2008. Fast solvers and efficient
  implementations for distance metric learning. In: ICML.

\bibitem[{Wu et~al.(2011)Wu, Mukunoki, Funatomi, Minoh, and
  Lao}]{Wu2011Optimizing}
Wu, Y., Mukunoki, M., Funatomi, T., Minoh, M., Lao, S., 2011. Optimizing mean
  reciprocal rank for person re-identification. In: Advanced Video and
  Signal-Based Surveillance.

\bibitem[{Xiong et~al.(2014)Xiong, Gou, Camps, and Sznaier}]{Xiong2014Person}
Xiong, F., Gou, M., Camps, O., Sznaier, M., 2014. Person re-identification
  using kernel-based metric learning methods. In: Proc. Eur. Conf. Comp. Vis.

\bibitem[{Yang et~al.(2012)Yang, Li, Mahdavi, Jin, and Zhou}]{nips2012nystrom}
Yang, T., Li, Y.-F., Mahdavi, M., Jin, R., Zhou, Z.-H., 2012. Nystr\"{o}m
  method vs random fourier features: a theoretical and empirical comparison.
  In: Proc. Adv. Neural Inf. Process. Syst.

\bibitem[{Zhao et~al.(2013{\natexlab{a}})Zhao, Ouyang, and
  Wang}]{Zhao2013SalMatch}
Zhao, R., Ouyang, W., Wang, X., 2013{\natexlab{a}}. Person re-identification by
  salience matching. In: Proc. IEEE Int. Conf. Comp. Vis.

\bibitem[{Zhao et~al.(2013{\natexlab{b}})Zhao, Ouyang, and
  Wang}]{Zhao2013Unsupervised}
Zhao, R., Ouyang, W., Wang, X., 2013{\natexlab{b}}. Unsupervised salience
  learning for person re-identification. In: Proc. IEEE Conf. Comp. Vis. Patt.
  Recogn.

\bibitem[{Zhao et~al.(2014)Zhao, Ouyang, and Wang}]{Zhao2014Learning}
Zhao, R., Ouyang, W., Wang, X., 2014. Learning mid-level filters for person
  re-identfiation. In: Proc. IEEE Conf. Comp. Vis. Patt. Recogn.

\bibitem[{Zheng et~al.(2013)Zheng, Gong, and Xiang}]{Zheng2013PAMI}
Zheng, W., Gong, S., Xiang, T., 2013. Re-identification by relative distance
  comparision. {IEEE} Trans. Pattern Anal. Mach. Intell. 35~(653-668), 3.

\bibitem[{Zheng et~al.(2009)Zheng, Gong, and Xiang}]{Zheng2009Associating}
Zheng, W.-S., Gong, S., Xiang, T., 2009. Associating groups of people. In:
  Proc. British Mach. Vis. Conf.

\bibitem[{Zheng et~al.(2011)Zheng, Gong, and Xiang}]{Zheng2011Person}
Zheng, W.-S., Gong, S., Xiang, T., 2011. Person re-identification by
  probabilistic relative distance comparison. In: Proc. IEEE Conf. Comp. Vis.
  Patt. Recogn.

\end{thebibliography}

\end{document}